\documentclass[manuscript,screen]{acmart}

\usepackage{subfigure}
\usepackage{amsmath}
\DeclareMathOperator{\E}{\mathbb{E}}

\newcommand{\argmin}{\mathop{\rm argmin}\limits}

\usepackage[linesnumbered,ruled,vlined]{algorithm2e} 

\AtBeginDocument{%
  \providecommand\BibTeX{{%
    \normalfont B\kern-0.5em{\scshape i\kern-0.25em b}\kern-0.8em\TeX}}}

\copyrightyear{2020} 
\acmYear{2020} 
\setcopyright{acmlicensed}\acmConference[RecSys '20]{Fourteenth ACM Conference on Recommender Systems}{September 22--26, 2020}{Virtual Event, Brazil}
\acmBooktitle{Fourteenth ACM Conference on Recommender Systems (RecSys '20), September 22--26, 2020, Virtual Event, Brazil}
\acmPrice{15.00}
\acmDOI{10.1145/3383313.3412261}
\acmISBN{978-1-4503-7583-2/20/09}

\begin{document}

\title{Unbiased Learning for the Causal Effect of Recommendation}

\author{Masahiro Sato}
\email{sato.masahiro@fujixerox.co.jp}
\affiliation{%
	\institution{Fuji Xerox}
	\city{Yokohama}
	\country{Japan}
}

\author{Sho Takemori}
\email{takemori.sho@fujixerox.co.jp}
\affiliation{%
	\institution{Fuji Xerox}
	\city{Yokohama}
	\country{Japan}
}

\author{Janmajay Singh}
\email{janmajay.singh@fujixerox.co.jp}
\affiliation{%
	\institution{Fuji Xerox}
	\city{Yokohama}
	\country{Japan}
}

\author{Tomoko Ohkuma}
\email{ohkuma.tomoko@fujixerox.co.jp}
\affiliation{%
	\institution{Fuji Xerox}
	\city{Yokohama}
	\country{Japan}
}

\renewcommand{\shortauthors}{M. Sato, et al.}

\begin{abstract}
	Increasing users' positive interactions, such as purchases or clicks, is an important objective of recommender systems.
	Recommenders typically aim to select items that users will interact with.
	If the recommended items are purchased, an increase in sales is expected.
	However, the items could have been purchased even without recommendation.
	Thus, we want to recommend items that results in purchases caused by recommendation.
	This can be formulated as a ranking problem in terms of the causal effect.
	Despite its importance, this problem has not been well explored in the related research.
	It is challenging because the ground truth of causal effect is unobservable, and estimating the causal effect is prone to the bias arising from currently deployed recommenders.
	This paper proposes an unbiased learning framework for the causal effect of recommendation.
	Based on the inverse propensity scoring technique, the proposed framework first constructs unbiased estimators for ranking metrics.
	Then, it conducts empirical risk minimization on the estimators with propensity capping, which reduces variance under finite training samples.
	Based on the framework, we develop an unbiased learning method for the causal effect extension of a ranking metric.
	We theoretically analyze the unbiasedness of the proposed method and empirically demonstrate that the proposed method outperforms other biased learning methods in various settings.
\end{abstract}

\ccsdesc[500]{Information systems~Recommender systems}
\ccsdesc[500]{Computing methodologies~Learning from implicit feedback}

\keywords{causal inference, treatment effect, learning to rank}

\maketitle

\section{Introduction}
Recommender systems have been used in various services to increase sales and user engagement \cite{Jannach19}.
Hence, both industry and academia have long been pursuing better recommender models.
Sales and user engagement are closely associated with positive user interactions, such as purchases and clicks.
Most previous researches have focused on accurately predicting the interactions of users and recommending items that have a higher purchase or click probability.
However, even without recommendations, the recommended items might have been clicked on \cite{Sharma15}, and recommending such items would not increase positive interactions.

The change caused by a certain treatment (in our case, the treatment is the recommendation) is called the causal effect or treatment effect \cite{Imbens15,Hernan20}.
To increase sales and user engagement, causal effect should be the objective of recommenders.
This leads to a novel problem of ranking items by the causal effect.

This ranking problem is difficult due to the following two reasons.
First, the ground truth of the causal effect is not observable.
If a certain item is purchased only if it is recommended, then recommending the item has positive causal effect.
The outcomes for recommended and unrecommended cases are called potential outcomes \cite{Rubin74}, and
the causal effect is defined as the difference of these potential outcomes.
The causal effect cannot be directly observed since an item is either recommended or not for a specific user at a given time.
This unobservable nature is known as the {\it fundamental problem of causal inference} \cite{Holland86}.
Second, estimation of the causal effect is prone to suffer from the bias due to the confounding between a recommendation and potential outcomes.
If a recommended item is purchased and another non-recommended item is not purchased, the purchase might be attributed to the causal effect of recommendation.
However, the purchase might be because the recommended item highly matches preferences of the user, i.e., the item might have been purchased irrespective of whether it was recommended.
The causal effect is overestimated in this example.
If we train a recommender based on biased estimates of the ground truth, the trained model cannot generate an optimal ranking.

In this paper, we propose an unbiased learning method for the causal effect of recommendations. 
We first define the ranking metrics for the causal effect by extending metrics commonly used for observable feedbacks.
Then we derive an unbiased estimator of the metrics by using the inverse propensity scoring (IPS) \cite{Hirano03,Lunceford04} technique.
Based on the estimator, we construct an unbiased learning method that optimizes the causal effect extension of the average rank metric.
We derive our method with theoretical justification.
This gives a principal basis for further extensions to address the new ranking problem.
We conduct experiments with semi-synthetic datasets generated from real-world data.
The empirical results demonstrate that the proposed method outperforms biased learning methods and works robustly in a wide range of simulation conditions.

The contributions of this paper are summarized as follows.
\begin{itemize}
	\item We develop IPS-based unbiased estimators for evaluating ranking performance in terms of the causal effect of recommendations.
	\item We propose an unbiased learning method for the causal effect by the empirical risk minimization of the estimators.
	\item We provide theoretical analysis for unbiasedness of the estimators and for the variance under finite samples.
	\item We conduct experiments with semi-synthetic datasets and demonstrate the effectiveness of the proposed method.
\end{itemize}

\section{Related Work}
\subsection{Causal Effect of Recommendation}
In general, recommender systems have a positive effect on various business values such as sales and user engagement \cite{Jannach19}.
The average causal effect of recommendations has been measured through A/B testing \cite{Dias08, Lee14}, which compares sales volumes with and without recommender systems.
Another type of A/B testings compares click-through rates or conversion rates of recommended items by different recommenders.
However, this approach does not estimate the causal effect\footnote{It can be regarded the causal effect in some special cases, such as advertisement, where click or purchase is only possible if recommended (i.e., no user-initiated positive interactions).} since the items that are recommended and clicked could have been clicked even without recommendations \cite{Sharma15}.

Even when the desired goal is a positive causal effect, most recommenders are designed to recommend items that match the preferences of the users, or equivalently, items that have a high purchase probability.
These recommenders are based on the underlying assumption that the increase of probability that a user will buy an item is proportional to the probability that the user will buy this item in the absence of recommendations \cite{Shani05}.
Unfortunately, this assumption is not necessarily true \cite{Sato16}.
Instead of predicting the natural user behavior without recommendations, some works \cite{Bonner18, Wang18} focus on outcome prediction under recommendations. 
Recently, recommendation strategies targeting the causal effect have been proposed \cite{Bodapati08, Sato16, Sato19}.
One strategy \cite{Bodapati08, Sato16} builds a purchase prediction model that can predict probabilities with and without recommendations, and then ranks items by the difference of the probabilities.
Another strategy \cite{Sato19} tries to directly optimize ranking for the causal effect.
The latter has been shown to be more effective than the former.
However, the previous method with the latter strategy \cite{Sato19} could lead to heavily biased learning since it neglects the confounding between recommendation assignments and potential outcomes.
In this paper, we address this issue and build an unbiased learning method.
We also compare our method with the previous learning method in the experiment section.

\subsection{Unbiased Learning}
\begin{figure}[tp]
	\begin{center}
		\subfigure[Selection bias.]{\includegraphics[width=0.2\textwidth]{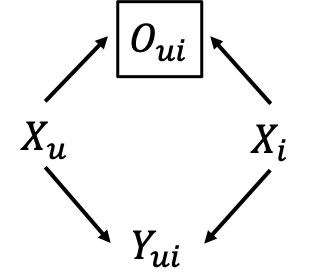}}
		\subfigure[Confounding.]{\includegraphics[width=0.2\textwidth]{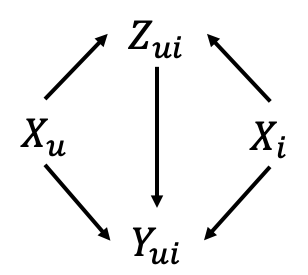}}
		\caption{
			Causal diagrams corresponding to previous unbiased learning (selection bias) and our work (confounding). 
			$X_{u}$ and $X_{i}$ are features of users and items, respectively, and $Y_{ui}$ is outcomes like purchases or clicks.
			$O_{ui}$ is the indicator of whether $Y_{ui}$ is observed and the rectangular signifies the observation is conditioned on $O_{ui}$.
			$Z_{ui}$ is the treatment that has causal effect on $Y_{ui}$. 
		}
	\label{fig:diagrams}
	\end{center}
\end{figure}

Unbiased learning has been intensively studied to solve the position bias in information retrieval \cite{Joachims17, Ai18, Agarwal19, Hu19} and the missing-not-at-randomness (MNAR) in feedbacks for recommendation \cite{Schnabel16, Wang19, Saito20}.
They first build unbiased estimators of targeted metrics by using IPS techniques \cite{Hirano03,Lunceford04} and then derive learning methods that achieve the empirical risk minimization (ERM) of the unbiased estimators.
These are cases of selection bias where observations of user feedbacks are conditioned on item exposure.
This work is similar to the previous unbiased learning in that both employ IPS techniques and ERM framework.
However, the bias we address is a case of confounding and there is a structural difference in bias mechanism as shown in Fig. \ref{fig:diagrams}.
Formerly, the selection bias is due to conditioning on common effects and the confounding is due to the presence of common causes (see chapter 8.3 in \cite{Hernan20}).
Further, while the ground truth is at least partially observable in previous works, it cannot be observed in our study, since we target the causal effect.
This is the first work to address the confounding bias that arises when we target the causal effect of recommendation. 

\section{Learning for the Causal Effect of Recommendation}
\begin{table*}[tp]
	\small
	\caption{Notations.}
	\label{tab:notation}
	\begin{tabular}{cc}
		\toprule
		Symbol & Description\\
		\midrule
		$Z_{ui}$ & Indicator variable of the treatment for user $u$ and item $i$. \\
		$Y_{ui}^\textnormal{T}$, $Y_{ui}^\textnormal{C}$ & 
		\begin{tabular}{c} Potential outcomes for user $u$ and item $i$ if recommended (treatment condition, $Z_{ui} = 1$),\\ and if NOT recommended (control condition, $Z_{ui} = 0$), respectively.\\
		\end{tabular}\\
		$\tau_{ui}$ & Causal effect of recommending item $i$ to user $u$, defined as $Y_{ui}^\textnormal{T} - Y_{ui}^\textnormal{C}$.\\
		$Y_{ui}$ & Observed outcome for user $u$ and item $i$, expressed as $Z_{ui} Y_{ui}^\textnormal{T} + (1-Z_{ui}) Y_{ui}^\textnormal{C}$.\\
		$P_{ui}$ & Probability that item $i$ is recommended to user $u$. (It is also called \textit{propensity}.)\\
		$X_{u}$, $X_{i}$ &  Features of user $u$ and item $i$, respectively.\\
		\midrule
		$\hat{s}_{ui}$ & Score predicted by a certain scoring function $f_M (X_{u}, X_{i})$ of model $M$.\\
		$\lambda_{ui}$ & Weighting function that depends on the ranking position $r (\hat{s}_{ui})$ of item $i$ for user $u$.\\
		$\Delta_{u}$, $\Delta_{u}^\textnormal{IPS}$, $\Delta_{u}^\textnormal{CIPS}$ & The ranking metric for user $u$, and its IPS and capped IPS estimates, respectively.\\
		$R(M)$, $R^\textnormal{IPS}(M)$, $R^\textnormal{CIPS}(M)$ & The average of ranking metric, and its IPS and capped IPS estimates, respectively.\\
		\bottomrule
	\end{tabular}
\end{table*}

\subsection{Definition of the Causal Effect}
Consider observations of the interactions of users on items, such as purchases or clicks.
Let $Y_{ui} \in \{0, 1\}$ denote an interaction of user $u \in \{1,...,U\}$ on item $i \in \{1,...,I\}$, and $Z_{ui} \in \{0, 1\}$ denote the binary indicator for the recommendation (also called treatment assignment).
Let the potential outcomes of the interactions be $Y_{ui}^\textnormal{T}$ and $Y_{ui}^\textnormal{C} \in \{0, 1\}$ when item $i$ is recommended to $u$ ($Z_{ui} = 1$) and when it is not recommended ($Z_{ui} = 0$), respectively.
$Y_{ui}$ is expressed as,
\begin{equation}
\label{eq:observed_outcome}
Y_{ui} = Z_{ui} Y_{ui}^\textnormal{T} + (1-Z_{ui}) Y_{ui}^\textnormal{C}.
\end{equation}
The causal effect $\tau_{ui}$ of recommending item $i$ to user $u$ is defined as the difference of the two potential outcomes \cite{Rubin74},
\begin{equation}
\tau_{ui} = Y_{ui}^\textnormal{T} - Y_{ui}^\textnormal{C},
\end{equation}
that takes ternary values, $\tau_{ui} \in \{-1, 0, 1\}$.
Note that either $Y_{ui}^\textnormal{T}$ or $Y_{ui}^\textnormal{C}$ can be observed at specific time, hence $\tau_{ui}$ is not directly observable.
Let $X_u$ and $X_i$ be the features of users and items, respectively, that affect treatment assignment $Z_{ui}$ and potential outcomes $Y_{ui}^\textnormal{T}$ and $Y_{ui}^\textnormal{C}$.
The causal diagram is shown in Fig. \ref{fig:diagrams} (b).

\subsection{Ranking Metric for the Causal Effect}
\label{subsec:ranking_metric}
Typically, items are ranked by scores $\hat{s}_{ui}$, which are predicted by a certain scoring function $f_M$ of model $M$.
\begin{equation}
\hat{s}_{ui} = f_M(X_u, X_i).
\end{equation}
Items are sorted by this score and each item gets its ranking position, $r (\hat{s}_{ui})$.
Ranking performance metric for the binary feedback $Y_{ui}$ can be expressed as,
\begin{equation}
\label{eq:delta_u_y}
\Delta_u  = \frac{1}{I} \sum_i \lambda(r (\hat{s}_{ui})) Y_{ui},
\end{equation}
where $\lambda(r (\hat{s}_{ui}))$ is a weighting function that depends on the rank $r (\hat{s}_{ui})$ of item $i$ ordered by $\hat{s}_{ui}$.
For brevity, we define $\lambda_{ui} = \lambda(r (\hat{s}_{ui}))$.
Various popular metrics are expressed in the above general form \cite{Yang18,Agarwal19}.
For example,
\begin{equation}
\textnormal{Area under the Curve (AUC): } \lambda_{ui}^{\textnormal{AUC}} = \frac{(I - r (\hat{s}_{ui}))}{\sum_i \boldsymbol{1}(Y_{ui}=1) },
\end{equation}
\begin{equation}
\textnormal{Average Rank (AR): } \lambda_{ui}^{\textnormal{AR}} = -r (\hat{s}_{ui}),
\end{equation}
\begin{equation}
\textnormal{Precision@\textit{k} (P@\textit{k}): } \lambda_{ui}^{\textnormal{P@\textit{k}}} = \boldsymbol{1}(r (\hat{s}_{ui}) <= k) I/k,
\end{equation}
\begin{equation}
\textnormal{Discounted Cumulative Gain (DCG): } \lambda_{ui}^{\textnormal{DCG}} = I/\log_2 (1+r (\hat{s}_{ui})),
\end{equation}
where $\boldsymbol{1}()$ is an indicator function.
Negative sign is added for the AR to make it a reward metric.
By neglecting a constant term $I$ in $\lambda_{ui}^{\textnormal{AUC}}$, both $\lambda_{ui}^{\textnormal{AUC}}$ and $\lambda_{ui}^{\textnormal{AR}}$ is proportional to the negative rank, and AUC and AR are similar metrics.

To maximize the impact of recommendations, items with positive causal effect (not just the ones with positive interaction) need to be recommended.
Hence we substitute $Y_{ui}$ with $\tau_{ui}$ in Eq. (\ref{eq:delta_u_y}) and construct the ranking metrics for the causal effect.
\begin{equation}
\label{eq:delta_u_tau}
\Delta_u  = \frac{1}{I} \sum_i \lambda_{ui} \tau_{ui}.
\end{equation}
To clarify that the metrics are extensions to the causal effect, we call them as causal average rank (CAR), causal precision @\textit{k} (CP@\textit{k}), and causal discounted cumulative gain (CDCG).
Note that CP@\textit{k} equals the average causal effect of top-\textit{k} items, which is also called uplift@\textit{k} in \cite{Sato19}.

Let $R(M)$ be the average of the ranking metric over all users.
\begin{equation}
R(M) = \frac{1}{U} \sum_u \Delta_u = \frac{1}{UI} \sum_u \sum_i \lambda_{ui} \tau_{ui}.
\end{equation}
Since model $M$ is evaluated by $R(M)$, the goal of learning is to maximize $R(M)$.
If $R(M)$ is observable, this can be achieved by empirical risk minimization (ERM) of $R(M)$.
\begin{equation}
\hat{M} = \argmin_{M \in \mathcal{H}_M} (-R(M)),
\end{equation}
where $\mathcal{H}_M$ is a hypothesis space of models.
However, this ERM is not feasible since $R(M)$ comprises unobservable $\tau_{ui}$.
In the next section, we construct an unbiased estimator of learning objective.

\section{Unbiased Learning for Causal Effect}
\subsection{Unbiased Estimator}
\label{subsec:unbiased_estimator}
Since the causal effect is not directly observable, its estimate is needed.
Naively, the average causal effect over whole user-item pairs can be estimated as the difference between the averages of outcomes under treatment and control, $\sum Z_{ui} Y_{ui}/\sum Z_{ui} - \sum (1-Z_{ui}) Y_{ui}/\sum (1-Z_{ui})$.
This leads to the following estimate of individual causal effect.
\begin{equation}
\label{eq:tau_naive}
\tilde{\tau}_{ui}^{\textnormal{Naive}} = \frac{Z_{ui} Y_{ui}}{\sum Z_{ui}/UI} - \frac{(1-Z_{ui}) Y_{ui}}{\sum (1-Z_{ui})/UI}.
\end{equation}
However, this is a biased estimate when treatment and potential outcomes are confounded.
Instead, an unbiased estimate of the causal effect is defined as follows:
\begin{equation}
\label{eq:tau_IPS}
\tilde{\tau}_{ui}^{\textnormal{IPS}} = \frac{Z_{ui} Y_{ui}}{P_{ui}} - \frac{(1-Z_{ui}) Y_{ui}}{1-P_{ui}}.
\end{equation}
Here the observed outcomes $Y_{ui}$ are weighted by the inverse of recommendation probability $P_{ui}$ when recommended ($Z_{ui}=1$), and by the inverse of non-recommendation probability $1-P_{ui}$ when not recommended ($Z_{ui}=0$).
These weighting compensates for unevenness of recommendation assignment.
It is called inverse propensity score (IPS) weighting \cite{Lunceford04, Hirano03}.
From Eq. (\ref{eq:observed_outcome}),
\begin{align} 
\tilde{\tau}_{ui}^{\textnormal{IPS}}
&= \frac{Z_{ui} (Z_{ui} Y_{ui}^\textnormal{T} + (1-Z_{ui}) Y_{ui}^\textnormal{C})}{P_{ui}} - \frac{(1-Z_{ui}) (Z_{ui} Y_{ui}^\textnormal{T} + (1-Z_{ui}) Y_{ui}^\textnormal{C})}{1-P_{ui}}\nonumber\\
&= \frac{Z_{ui} Y_{ui}^\textnormal{T}}{P_{ui}} - \frac{(1-Z_{ui}) Y_{ui}^\textnormal{C}}{1-P_{ui}}.
\end{align}
This holds since $Z_{ui}(1-Z_{ui}) = 0$, $Z_{ui}^2 = Z_{ui}$, and $(1-Z_{ui})^2 = (1-Z_{ui})$.

Let define the IPS estimate of ranking metric, $\Delta_{u}^{\textnormal{IPS}}$, as,
\begin{equation}
\Delta_{u}^{\textnormal{IPS}} = \frac{1}{I} \sum_i \lambda_{ui} \tilde{\tau}_{ui}^{\textnormal{IPS}}.
\end{equation}
$\Delta_{u}^{\textnormal{IPS}}$ is unbiased since,
\begin{align} 
\label{eq:e_delta_u_IPS}
\E_{\{Z_{ui}\}} [\Delta_{u}^{\textnormal{IPS}}]
&= \E_{\{Z_{ui}\}} \left[ \frac{1}{I} \sum_i \lambda_{ui} \tilde{\tau}_{ui}^{\textnormal{IPS}} \right] \nonumber\\
&= \frac{1}{I} \sum_i \E_{\{Z_{ui}\}} \left[ \lambda_{ui} \left(\frac{Z_{ui} Y_{ui}^\textnormal{T}}{P_{ui}} - \frac{(1-Z_{ui}) Y_{ui}^\textnormal{C}}{1-P_{ui}} \right)\right]\nonumber\\
&= \frac{1}{I} \sum_i \lambda_{ui} \left(\frac{P_{ui} Y_{ui}^\textnormal{T}}{P_{ui}} - \frac{(1-P_{ui}) Y_{ui}^\textnormal{C}}{1-P_{ui}} \right) \nonumber\\
&= \frac{1}{I} \sum_i \lambda_{ui} (Y_{ui}^\textnormal{T} - Y_{ui}^\textnormal{C}) = \Delta_{u}
\end{align}
In the third equality, we used the conditional independence $(Y_{ui}^\textnormal{T}, Y_{ui}^\textnormal{C}) \perp Z_{ui} | P_{ui}$ \cite{Lunceford04} .
In other words, recommendation assignment and potential outcomes are independent if the recommendation probability is known.
The unbiasedness holds for any weighting functions $\lambda_{ui}$ introduced in Subsection \ref{subsec:ranking_metric}.
This unbiased estimator forms a basis for our learning method.

In this study, we assume that the propensity is recorded together with user interaction logs as in \cite{Lefortier16}.
In the cases where the true propensities are not provided, they need to be estimated \cite{Hirano03}.
Here we derive the bias under estimated propensities, which is also used to derive the bias by capping in the next subsection.
Let $\hat{\Delta}_{u}^{\textnormal{IPS}}$ denote the metric with estimated propensity $\hat{P}_{ui}$.
It is obtained by substituting true propensity $P_{ui}$ with $\hat{P}_{ui}$ in Eq. (\ref{eq:tau_IPS}).
By the derivation similar to Eq. (\ref{eq:e_delta_u_IPS}), we can obtain the expectation of $\hat{\Delta}_{u}^{\textnormal{IPS}}$ as,
\begin{equation}
\E_{\{Z_{ui}\}} [\hat{\Delta}_{u}^{\textnormal{IPS}}]
= \frac{1}{I} \sum_i \lambda_{ui} \left(\frac{P_{ui} Y_{ui}^\textnormal{T}}{\hat{P}_{ui}} - \frac{(1-P_{ui}) Y_{ui}^\textnormal{C}}{1-\hat{P}_{ui}} \right).
\end{equation}
The bias of the metric becomes,
\begin{align} 
\textnormal{Bias}(\hat{\Delta}_{u}^{\textnormal{IPS}})
&= \E_{\{Z_{ui}\}} [\hat{\Delta}_{u}^{\textnormal{IPS}} - \Delta_{u}] \nonumber\\ 
&= \frac{1}{I} \sum_i \lambda_{ui}  \left(\left(1 - \frac{P_{ui}}{\hat{P}_{ui}} \right)Y_{ui}^\textnormal{T} - \left(1 - \frac{1-P_{ui}}{1-\hat{P}_{ui}} \right)Y_{ui}^\textnormal{C} \right).
\label{eq:bias_estimated_propensity}
\end{align}
If the propensity is correctly specified, i.e., $\hat{P}_{ui} = P_{ui}$, then the bias becomes zero.
The naive estimate with Eq. (\ref{eq:tau_naive}) can be regarded as a special case where $\hat{P}_{ui}=\sum Z_{ui}/UI \approx \sum P_{ui}/UI$, that is, the estimated propensity equals the average propensity.
This is obviously biased unless a recommender recommends items randomly with the same probability.

\subsection{Analysis for Finite Samples and Propensity Capping}
\label{subsec:finite_analysis}
In the previous subsection, we derive the unbiased estimator for the ranking metrics.
The unbiasedness means that the estimator converges to true values in expectation.
However, the estimate from finite samples can deviate from true values.
This subsection analyzes the case with finite samples and introduce the propensity capping to reduce the variance of the estimate.

Hoeffding's inequality \cite{Hoeffding94} states that the following inequality holds for independent (not necessarily identically distributed) random variables $V_1, ..., V_n$ that take values in intervals of $d_1, ... d_n$,
\begin{equation}
P\left(\left| \sum V_k - \E \left[\sum V_k\right] \right| \geq \epsilon \right) \leq 2 \exp\left(\frac{-2\epsilon^2}{\sum d_k^2} \right).
\end{equation}
This inequality provides probability $\zeta$ that our learning objective $R^{\textnormal{IPS}}(M)$ can deviate from $R(M)$ by $\epsilon$ (refer to the supplementary materials of \cite{Schnabel16, Wang19} for similar derivation).
In our case, since $\tilde{\tau}_{ui}^{\textnormal{IPS}}$ is a random variable which takes values in intervals from $\frac{1}{P_{ui}}$ to $-\frac{1}{1-P_{ui}}$,
\begin{equation}
\label{eq:bound}
P\left( \left| R^{\textnormal{IPS}}(M) - R(M) \right| \geq \epsilon \right) \leq 2 \exp\left(\frac{-2\epsilon^2 U^2 I^2}{\sum d_{ui}^2} \right) = \zeta,
\end{equation}
where $d_{ui} = \lambda_{ui} \left(\frac{1}{P_{ui}} + \frac{1}{1-P_{ui}}\right)$.
Thus, at least with probability $1-\zeta$, the deviation of the IPS estimator is bounded as,
\begin{equation}
\label{eq:tail_bound_IPS}
\left| R^{\textnormal{IPS}}(M) - R(M) \right| \leq \frac{1}{UI} \sqrt{\frac{\log \frac{2}{\zeta}}{2}} \sqrt {\sum d_{ui}^2}.
\end{equation}
This is obtained by solving the relationship between $\zeta$ and $\epsilon$ in Eq. (\ref{eq:bound}).

The derived bound suggests that if $P_{ui}$ is close to 1 or 0, the IPS estimate $R^{\textnormal{IPS}}(M)$ can largely deviate from the real objective $R(M)$.
To remedy this, we cap the propensity to bound $1/P_{ui}$ and $1/(1 - P_{ui})$ in a proper range.
Using capping parameters $\chi^\textnormal{T}, \chi^\textnormal{C}$, we threshold the denominators of the first and second terms in Eq. (\ref{eq:tau_IPS}) as $\max(P_{ui}, \chi^\textnormal{T})$ and $\max(1-P_{ui}, \chi^\textnormal{C})$, respectively.
The propensity capping has been used for off-policy learning \cite{Bottou13} and unbiased learning under selection bias \cite{Saito20}.
The bias incurred by the capped inverse propensity scoring (CIPS) can be derived from Eq. (\ref{eq:bias_estimated_propensity}),
\begin{align} 
\textnormal{Bias}(R^{\textnormal{CIPS}}(M))
&= \frac{1}{UI} \sum_u \sum_i \lambda_{ui}  \left(\left(1 - \frac{P_{ui}}{\chi^\textnormal{T}} \right)Y_{ui}^\textnormal{T} \boldsymbol{1}(P_{ui} < \chi^\textnormal{T})- \left(1 - \frac{1-P_{ui}}{\chi^\textnormal{C}} \right)Y_{ui}^\textnormal{C} \boldsymbol{1}(P_{ui} > 1-\chi^\textnormal{C})\right).
\end{align}
With the expense of the above small bias, the capping tightens the bound (\ref{eq:tail_bound_IPS}).
\begin{equation} 
\left| R^{\textnormal{CIPS}}(M) - R(M) \right| \leq |\textnormal{Bias}(R^{\textnormal{CIPS}}(M))| + \frac{1}{UI} \sqrt{\frac{\log \frac{2}{\zeta}}{2}} 
\sqrt {\sum \lambda_{ui}^2 \left( \frac{1}{\max(P_{ui}, \chi^\textnormal{T})} + \frac{1}{\max(1-P_{ui}, \chi^\textnormal{C})} \right)^2}.
\end{equation}
We use this CIPS estimator for ERM.
\begin{equation}
\hat{M} = \argmin_{M \in \mathcal{H}_M} \left(-R^{\textnormal{CIPS}}(M) \right).
\end{equation}

\subsection{Unbiased Learning for Causal AR}
\label{subsec:unbiased_learning}
In this subsection, we derive an efficient optimization method for CAR, in which the weighing function is $\lambda_{ui} = -r(\hat{s}_{ui})$.
Let define the local loss as $\delta_{ui} = -\lambda_{ui} \tilde{\tau}_{ui}$, and CIPS estimate of potential outcomes as $\tilde{Y}_{ui}^\textnormal{T} = \frac{Z_{ui} Y_{ui}}{\max(P_{ui}, \chi^\textnormal{T})}$ and $\tilde{Y}_{ui}^\textnormal{C} = \frac{(1-Z_{ui}) Y_{ui}}{\max(1-P_{ui}, \chi^\textnormal{C})}$.
Since $r (\hat{s}_{ui}) = 1 + \sum_{j \neq i} \boldsymbol{1}(\hat{s}_{uj} > \hat{s}_{ui})$, the local loss is expressed as follows:
\begin{equation}
\label{eq:local_loss1}
\delta_{ui} 
= r (\hat{s}_{ui}) \left(\tilde{Y}_{ui}^\textnormal{T} - \tilde{Y}_{ui}^\textnormal{C} \right) \nonumber \\
= \left(1 + \sum_{j \neq i} \boldsymbol{1}(\hat{s}_{uj} > \hat{s}_{ui})\right) \left(\tilde{Y}_{ui}^\textnormal{T} - \tilde{Y}_{ui}^\textnormal{C} \right)
\end{equation}
The difficulty in directly optimizing the above metric lies in the non-differentiable indicator function.
A common way to tackle this challenge is optimizing either the differentiable approximation or differentiable upper bound \cite{Liu09}.
In the former approach, the indicator function can be approximated by the sigmoid function \cite{Qin10,Yan03}, $\boldsymbol{1}(x < 0) \approx 1/(1+\exp(\omega x))$, where $\omega$ is the parameter to adjust the steepness of the curve.
Let define $\hat{s}_{uij} = \hat{s}_{ui} - \hat{s}_{uj}$.
From Eq. (\ref{eq:local_loss1}), we can obtain the following approximation.
\begin{equation}
\label{eq:approximation}
\delta_{ui} \approx \left( \sum_{j \neq i} \frac{1}{1+\exp(\omega \hat{s}_{uij})} \right) \left( \tilde{Y}_{ui}^\textnormal{T} - \tilde{Y}_{ui}^\textnormal{C}\right) + const.
\end{equation}

In the second approach, the indicator function is upper bounded by the logistic loss, $\boldsymbol{1}(x < 0) \leq \log (1+\exp(-\omega x))$, or the hinge loss, $\boldsymbol{1}(x < 0) \leq \max(1-\omega x, 0)$.
These bounds have been successfully used for the optimization of AUC \cite{Rendle09} and that of generalized AUC \cite{Song15}.
We use the logistic loss upper bound in this paper.
To apply the upper bound, Eq. (\ref{eq:local_loss1}) is transformed using $I = 1 + \sum_{j \neq i} \boldsymbol{1}(\hat{s}_{uj} > \hat{s}_{ui}) + \sum_{j \neq i} \boldsymbol{1}(\hat{s}_{ui} > \hat{s}_{uj})$, and the following equation is obtained.
\begin{align} 
\label{eq:upper_bound}
\delta_{ui} 
&= r (\hat{s}_{ui}) \tilde{Y}_{ui}^\textnormal{T} +  (I - r (\hat{s}_{ui})) \tilde{Y}_{ui}^\textnormal{C} - I \tilde{Y}_{ui}^\textnormal{C} \nonumber \\
&= \left(1 + \sum_{j \neq i} \boldsymbol{1}(\hat{s}_{uj} > \hat{s}_{ui})\right) \tilde{Y}_{ui}^\textnormal{T} + \left(\sum_{j \neq i} \boldsymbol{1}(\hat{s}_{ui} > \hat{s}_{uj})\right) \tilde{Y}_{ui}^\textnormal{C} - I \tilde{Y}_{ui}^\textnormal{C}\nonumber \\
& \leq \left(\sum_{j \neq i} \log (1+\exp(-\omega \hat{s}_{uij})) \right) \tilde{Y}_{ui}^\textnormal{T} + \left(\sum_{j \neq i} \log (1+\exp(\omega \hat{s}_{uij})) \right) \tilde{Y}_{ui}^\textnormal{C} + const.
\end{align}

Both the approximation (Eq. (\ref{eq:approximation})) and upper bound (Eq. (\ref{eq:upper_bound})) take the form of summation over triplet $(u, i, j)$.
We can efficiently optimize recommender models by stochastic gradient descent (SGD) with the derivatives of the losses below.
\begin{equation} 
\label{eq:approximation_local_loss}
\delta_{uij}^{AP} = \frac{Z_{ui} Y_{ui}}{\max(P_{ui}, \chi^\textnormal{T})}\frac{1}{1+\exp(\omega \hat{s}_{uij})} - \frac{(1-Z_{ui}) Y_{ui}}{\max(1-P_{ui}, \chi^\textnormal{C})}\frac{1}{1+\exp(\omega \hat{s}_{uij})}.
\end{equation}
\begin{equation} 
\label{eq:upper_bound_local_loss}
\delta_{uij}^{UB} = \frac{Z_{ui} Y_{ui}}{\max(P_{ui}, \chi^\textnormal{T})}\log (1+\exp(-\omega \hat{s}_{uij})) + \frac{(1-Z_{ui}) Y_{ui}}{\max(1-P_{ui}, \chi^\textnormal{C})}\log (1+\exp(\omega \hat{s}_{uij})).
\end{equation}
Note that if item $i$ is not purchased by user $u$ (i.e., $Y_{ui}=0$), then the above loss disappears.
Hence, $(u, i)$ is sampled from the user-item pairs with positive interactions.
Further, depending on $Z_{ui} \in \{0, 1\}$, either first or second terms remain for each training sample.

The pseudo code of the proposed learning method with the upper bound loss in Eq. (\ref{eq:upper_bound_local_loss}) is shown in Algorithm \ref{algo:DLCE}.
The algorithm for the approximation loss in Eq. (\ref{eq:approximation_local_loss}) can be obtained by substituting lines 6 and 8 with the correspondences.
Hyper-parameters of our method are learning rate $\eta$, regularization coefficient $\gamma$, the steepness of the surrogate losses $\omega$, and capping thresholds $\chi^\textnormal{T}$ and $\chi^\textnormal{C}$.
In Algorithm \ref{algo:DLCE}, $D = \{(u,i) |Y_{ui} > 0 \}$ is the logs of positive user interactions such as purchases or clicks.
In this work, we use a common matrix factorization (MF) model \cite{Koren09} for the scoring function $f_M$: $\hat{s}_{ui} = p_u \cdot q_i$,
where $p_u$ and $q_i$ are user and item latent factors, respectively.
Hence trainable parameters are $\Theta = \{p_u, q_i | u \in (1,..,U), i \in (1,...,I) \}$. 

\begin{algorithm}[tp] 
	\caption{Debiased Learning for the Causal Effect ({\textit{DLCE}}) with upper bound loss.}
	\label{algo:DLCE}
	\DontPrintSemicolon
	\KwIn{$\eta$, $\gamma$, $\omega$, $\chi^\textnormal{T}$, $\chi^\textnormal{C}$}
	\KwOut{$\Theta$}
	Random initialization of $\Theta$\;
	\While{not converged}{
		draw $(u, i)$ from $D$\;
		draw $j \neq i$ from $(1,...,I)$\;
		\eIf{$Z_{ui} = 1$}{
		$\Theta \gets \Theta + \eta \left(-\frac{1}{\max(P_{ui}, \chi^\textnormal{T})} \frac{\partial}{\partial \Theta} \log (1+\exp(\omega \hat{s}_{uij})) - \gamma \Theta \right)$\;
		}{
		$\Theta \gets \Theta + \eta \left(-\frac{1}{\max(1-P_{ui}, \chi^\textnormal{C})} \frac{\partial}{\partial \Theta} \log (1+\exp(-\omega \hat{s}_{uij})) -  \gamma \Theta \right)$\;
		}
	}
	\Return{$\Theta$}\;
\end{algorithm}

\subsection{Practical Applicability} 
The proposed method is a generic pairwise learning method and can be used to modify a wide class of recommendation models that leverage auxiliary information or neural architectures.
This is comparable to models which augment and improve upon BPR \cite{Rendle10,He16,Hidasi16}.
The practitioner need to collect recommendation logs ($Z_{ui}$) and propensities ($P_{ui}$) in addition to commonly used implicit feedback ($Y_{ui}$).
If the propensities are not available, estimated propensities\footnote{Previous work \cite{Sato19} describes the propensity estimation in the experiment section.} can be used instead \cite{Hirano03}.
Using these data, the training loss of either Eq. (\ref{eq:upper_bound_local_loss}) or (\ref{eq:approximation_local_loss}) should be applied for training the model. 
When we apply it online, we can expect an increase in total user interactions, such as purchases or clicks.
Note that interactions only for recommended items might decrease, since our method aims for the causal effect and avoids recommending items that would be purchased whether recommended or not.

\section{Empirical Evaluation}

\subsection{Experiment Setup\protect\footnote{The codes are available on arXiv as ancillary files.}}

\subsubsection{Generation of Semi-synthetic Datasets}

We generated semi-synthetic datasets based on publicly available Dunnhumby dataset\footnote{https://www.dunnhumby.com/careers/engineering/sourcefiles}.
Semi-synthetic data are required for two reasons.
First, there is no ground truth in real data because of the unobservable nature of causal effect.
Although the proposed learning method only uses observable variables, the ground truth is required for evaluation.
Secondly, the condition of recommendation assignment needs to be adjusted to see how the unevenness of propensity distribution affects the performance.
The datasets were generated by the following steps.
\begin{enumerate}
	\item \textbf{Preprocessing the base dataset}.
	The Dunnhumby dataset includes purchase and promotion logs of 2,500 users at a retailer for 93 weeks.
	Following \cite{Sato19}, the items featured in the weekly mailer are considered as recommendations.
	To ensure reliable estimate of purchase probabilities in the next step, we filtered the dataset according to the following conditions: users with at least 10 weeks of purchase logs, items with at least 10 weeks of purchase logs, and items with both treatment and control conditions.
	Two variants are prepared in terms of item granularity: product categories and product ids as items. 
	By applying the above procedure, we obtained the observations $\{Y_{uit}, Z_{uit}\}$, where $t$ denotes the $t$-th week.
	Moreover, we introduce the variable $V_{ut} = \boldsymbol{1}( \sum_i Y_{uit} > 0)$, which indicates that user $u$ purchased at least one item, hence visited the retailer, at the $t$-th week.

	\item \textbf{Modeling purchase probabilities}.
	We calculated the purchase probabilities for each user-item pair with and without recommendation by averaging the observations of whole 93 weeks. 
	\begin{equation}
	\mu_{ui}^\textnormal{T} = P(Y_{ui}^\textnormal{T}=1) = a_{ui}^\textnormal{T}/b_{ui}^\textnormal{T},\; 
	\mu_{ui}^\textnormal{C} = P(Y_{ui}^\textnormal{C}=1) = a_{ui}^\textnormal{C}/b_{ui}^\textnormal{C},
	\end{equation}
	where $a_{ui}^\textnormal{T}=\sum_t Z_{uit} Y_{uit}$, $b_{ui}^\textnormal{T}=\sum_t Z_{uit} V_{ut}$, $a_{ui}^\textnormal{C}=\sum_t (1-Z_{uit}) Y_{uit}$, and $b_{ui}^\textnormal{C}=\sum_t (1-Z_{uit})V_{ut}$.
	To account for uncertainty of the above estimate, we further add Beta prior, based on the averages over all users.
	\begin{equation}
	\mu_{ui}^\textnormal{T} = \frac{a_{ui}^\textnormal{T} + w a_{*i}^\textnormal{T}}{b_{ui}^\textnormal{T} + w b_{*i}^\textnormal{T}}, \quad  \mu_{ui}^\textnormal{C} = \frac{a_{ui}^\textnormal{C} + w a_{*i}^\textnormal{C}}{b_{ui}^\textnormal{C} + w b_{*i}^\textnormal{C}},
	\end{equation}
	where $a_{*i}^\textnormal{T}=\sum_u a_{ui}^\textnormal{T}/U$, $b_{*i}^\textnormal{T}=\sum_u b_{ui}^\textnormal{T}/U$, $a_{*i}^\textnormal{C}=\sum_u a_{ui}^\textnormal{C}/U$, and $b_{*i}^\textnormal{C}=\sum_u b_{ui}^\textnormal{C}/U$.
	The weight of the prior $w$ was tuned for the proper prediction of unseen week.
	More specifically, we first obtained the probabilities from first 92 weeks, and then measured the accuracy for the prediction of the last week.
	The accuracy was measured by Brier score \cite{Brier50, Snoek19}, which is the squared error of the predicted probabilities.
	We explored $w \in \{0.1, 0.2, .., 2.0\}$ and chose $w$ that achieved the lowest Brier score (0.4 and 0.9 for category-level and item-level granularity, respectively).

	\item \textbf{Modeling propensities}.
	We modeled the propensities in two ways: the first one is based on the original dataset, the second one simulates recommendations personalized to the preferences of the users.
	For the original setting, we calculate propensities based on the statistics of the original dataset.
	\begin{equation} 
	\begin{split}
	P_{ui} = \frac{a_{ui}^Z + w a_{*i}^Z}{b_{ui}^Z + w b_{*i}^Z}, \quad
	a_{ui}^Z=\sum_t Z_{uit}, \quad
	b_{ui}^Z=\sum_t V_{ut}, \\
	a_{*i}^Z=\sum_u a_{ui}^Z/U, \quad
	b_{*i}^Z=\sum_u b_{ui}^Z/U.
	\end{split}
	\end{equation}
	For the personalized setting\footnote{In the previous version of this paper, there is a mistake in the generation of datasets for the personalized setting. The propensity distribution was skewed as intended but the propensity did not properly depend on the personalized ranking. We have debugged it and conducted experiments again for the personalized settings.}, we first rank items for each user by the purchase probabilities, $\mu_{ui}=P(Y_{ui}=1)$, in the original setting.
	Then set the propensities as, 
	\begin{equation}
	\label{eq:gen_propensity}
	P_{ui} = \min \left(1, \alpha \left( 1/\textnormal{rank}\right)^\beta \right).
	\end{equation}
	The parameters $\alpha$ and $\beta$ allow control over the average and the unevenness of propensities, respectively.
	$\alpha$ is adjusted so as to make the average propensity equal to that of the original setting.
	$\beta$ is set to 2.0 for the default condition.
	Experiments with various $\beta$ values are described in Subsection \ref{subsec:unevenness_propensity}.
	The personalized setting is to simulate a common situation where a currently running recommender tends to select items that match the preference of the users with higher probabilities.
	For both the original and personalized settings, the propensities are clipped in the range $[10^{-6}, 1-10^{-6}]$ after the above procedure.
	
	\item \textbf{Generation of observed data}.
	We sampled the potential outcomes and recommendation assignments by using Bernoulli distributions.
	\begin{equation}
	Y_{ui}^\textnormal{T} \sim \textnormal{Bernoulli}(\mu_{ui}^\textnormal{T}), \;
	Y_{ui}^\textnormal{C} \sim \textnormal{Bernoulli}(\mu_{ui}^\textnormal{C}), \;
	Z_{ui} \sim \textnormal{Bernoulli}(P_{ui}).
	\end{equation}
	Then, the causal effect $\tau_{ui}$ and observed outcomes $Y_{ui}$ can be obtained as,
	\begin{equation}
	\tau_{ui} = Y_{ui}^\textnormal{T} - Y_{ui}^\textnormal{C}, \quad
	Y_{ui} = Z_{ui} Y_{ui}^\textnormal{T} +(1-Z_{ui}) Y_{ui}^\textnormal{C}.
	\end{equation}
	Note that only $Y_{ui}, Z_{ui}$, and $P_{ui}$ are observable and hence used for training, while $\tau_{ui}$ is provided for evaluation.
	This sampling is repeated $n$ times to generate dataset for $n$ weeks.
	Training, validation, and test datasets are independently sampled for $n_{train}=10$, $n_{validation}=1$, and $n_{test}=10$ times, respectively.
\end{enumerate}

\begin{figure}[htbp]
	\centering
	\setlength{\tabcolsep}{3pt} 
	\begin{minipage}{0.48\textwidth}
		\centering
		\captionsetup{type=table} 
		\begin{tabular}{cccccc}
			\toprule
			Granularity & \#User & \#Item & $\overline{\mu_{ui}^\textnormal{T}}$ & $\overline{\mu_{ui}^\textnormal{C}}$ & $\overline{P_{ui}}$ \\
			\midrule
			Category & 2,309 & 1,372 & 0.0145 & 0.0101 & 0.153\\
			Product & 2,290 & 11,331 & 0.0027 & 0.0010 & 0.088\\
			\bottomrule
		\end{tabular}
		\caption{Statistics of datasets.}
		\label{tab:stat_data}
	\end{minipage}
	\begin{minipage}{0.5\textwidth}
		\centering
		\subfigure[Category.]{\includegraphics[width=0.48\textwidth]{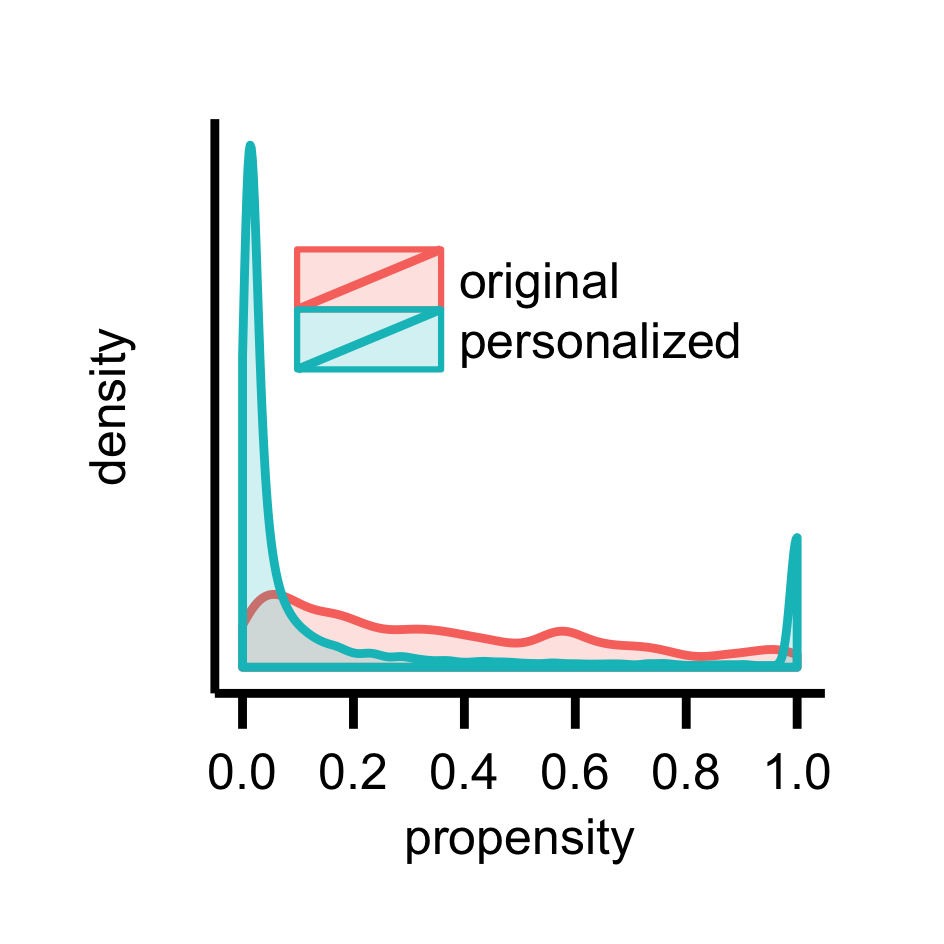}}
		\subfigure[Product.]{\includegraphics[width=0.48\textwidth]{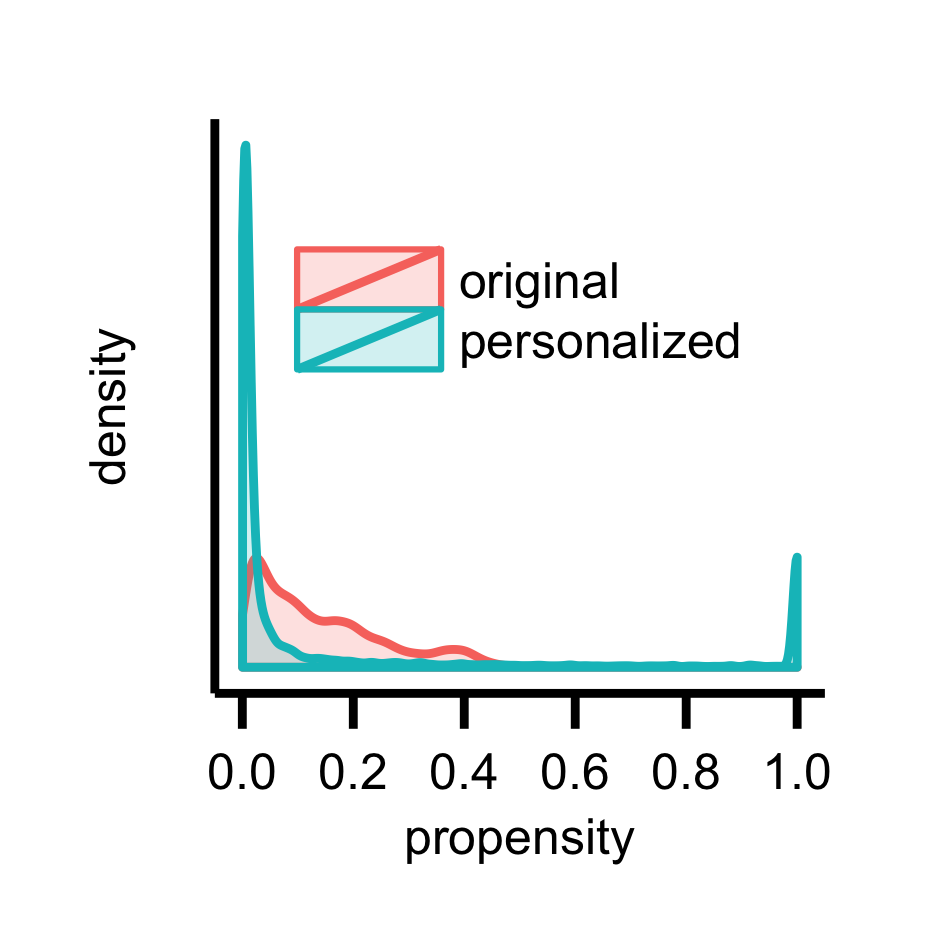}}
		\caption{Density distributions of the propensity.}
		\label{fig:propensity_dist}
	\end{minipage}
\end{figure}

Table \ref{tab:stat_data} shows the statistics of datasets and Fig. \ref{fig:propensity_dist} shows the density distributions of propensity for $D = \{(u,i) |Y_{ui} > 0 \}$.
The numbers of users and items are rather small. 
We can generate larger datasets if we use a larger base dataset such as Xing.
However, Xing dataset is no longer publicly available.
Another way is to generate fully synthetic datasets by arbitrary parameterizing $\mu_{ui}^\textnormal{T}$ and $\mu_{ui}^\textnormal{C}$, but it might not reflect reality.
Thus, we generate semi-synthetic datasets based on Dunnhumby.

\subsubsection{Compared Methods}
\label{subsubsec:compared_methods}
All the following methods except Random and Pop are compared by training the MF model.\\
\textbf{Random}: Items are ranked randomly.\\
\textbf{Pop}: Items are ranked by the global popularity, i.e., number of purchases, $\sum_u Y_{ui}$.\\
\textbf{BPR} \cite{Rendle09} : A commonly used pairwise learning method optimized for AUC of observed outcomes $Y_{ui}$. \\
\textbf{ULBPR} \cite{Sato19}: A pairwise learning method for the causal effect of recommendation. It targets for $\tau_{ui}$, but is a biased learning. \\
\textbf{ULRMF} \cite{Sato19}: A pointwise version of ULBPR.\\ 
\textbf{CausE} \cite{Bonner18}: A joint training of outcome prediction models with and without recommendations. It can be used for recommendations targeting the causal effect as demonstrated in \cite{Sato19}. We used the CausE-prod version.\\
\textbf{BLCE}: A biased learning for the causal effect $\tau_{ui}$ using the naive estimate defined in Eq. (\ref{eq:tau_naive}).\\
\textbf{DLTO}: An unbiased learning for the treated outcome $Y_{ui}^\textnormal{T}$, training only with cases where $Z_{ui}=1$.
(Line 8 in Algorithm \ref{algo:DLCE} is skipped.)\\
\textbf{DLCE}: The proposed unbiased learning for the causal effect $\tau_{ui}$.

Among the previous methods, ULBPR is an important baseline since it is a pairwise learning for the causal effect similar to DLCE. 
By comparing DLCE with ULBPR and BLCE, the impact of debiasing can be verified as these methods are prone to the confounding bias.
By comparing DLCE with DLTO, the impact of targeting causal effect can be verified as DLTO optimizes for the treated outcome $Y_{ui}^\textnormal{T}$.
In the preliminary experiment, both the approximation loss and the upper bound loss result in almost same performance, and the results for the upper bound loss are reported here.

The dimension size of MF was set to 200 for the original setting of propensity and 400 for the personalized setting. 
The performances of the compared methods saturate at these dimensional sizes. 
The steepness of curve $\omega$ is set to 1.0.
We use the same capping thresholds for treatment and control ($\chi = \chi^\textnormal{T} = \chi^\textnormal{C}$).
The regularization coefficient $\gamma$ and the capping threshold $\chi$, are tuned with the exploration ranges $\gamma \in \{0.3, 0.1, 0.03, 0.01, 0.003, 0.001, 0.0003, 0.0001\}$ and $\chi \in \{0.7, 0.5, 0.3, 0.1, 0.03, 0.01, 0.003, 0.001\}$, respectively. 

\subsubsection{Evaluation Protocol}
The methods are evaluated\footnote{The regular metrics such as P@10 do not correlate with their causal versions as described in \cite{Sato19}. We omit them for the space constraint.} with CP@10, CP@100, and CDCG that are defined in Subsection \ref{subsec:ranking_metric}.
As mentioned in the previous subsection, test datasets are generated for $n_{test}=10$ times.
We first measure the performance for each time and take average of 10 times for the final result.
We also report the standard deviation of the obtained performance.
Hyper-parameters of each method are tuned to optimize each metric with the validation dataset.
Note that the above metrics cannot be observed directly in the real-world, and hyper-parameters need to be tuned with the estimates of metrics.
However, in this work, we tune the hyper-parameters with the ground-truth metrics to focus on the comparison of each method\footnote{If we use the estimates of metrics for hyper-parameter tuning, the performance are affected both from the learning method itself and the estimator. Further the goodness of fit of the estimator might depend on the evaluated method, and this could prohibit the fair comparison.}.
Instead, we investigate the accuracy of the estimators separately in Subsection \ref{subsec:reliability_unbiased_estimators}.

\subsection{Comparison of DLCE to baselines}
The ranking performance of DLCE was compared with other baselines.
Tables \ref{tab:comparison_category} and \ref{tab:comparison_product} show the results.
Note that the ground truths are the same for the original and personalized setting and only observable training data are different.
Therefore, the random baseline performs in the same way in both settings.
DLCE improves upon BLCE and ULBPR, implying that the proposed IPS-based ERM framework can achieve expected unbiased learning.
DLCE outperforms DLTO in the original settings, but DLCE is outperformed by DLTO in the personalized settings.
This counter-intuitively suggests that optimizing a model for AR (conducted by DLTO) leads to better performance on CP and CDCG than optimizing for CAR (conducted by DLCE).
We further investigated the performance in CAR, which is the metric that DLCE and BLCE directly optimize for.
As shown in Table \ref{tab:comparison_car}, DLCE performs better than DLTO in CAR.

Regarding other baselines, ULRMF, a pointwise counterpart of ULBPR, is superior to ULBPR in the ogirinal settings, and is inferior to ULBPR in the personalized settings.
CausE is better than ULRMF and ULBPR in category-level granularity (Table \ref{tab:comparison_category}) and ULRMF or ULBPR tend to be better than CausE in product-level granularity (Table \ref{tab:comparison_product}).
Our DLCE mostly outperforms these baselines and the superiority is more prominent in product-level granularity that has larger number of items.
Hence, we can expect that our method is effective for even larger datasets.

\begin{table*}[htbp]
	\small
	\caption{Performance comparison in category-level granularity of items.
		The best results are highlighted in bold.
	}
	\label{tab:comparison_category}
	\centering
	\begin{tabular}{lcccccc}
		\toprule
		& \multicolumn{3}{c}{Category-Original} & \multicolumn{3}{c}{Category-Personalized ($\beta = 2.0$)}\\
		\cmidrule(lr){2-4} \cmidrule(lr){5-7} 
		& CP@10 & CP@100 & CDCG & CP@10 & CP@100 & CDCG \\
		\midrule
		Random & $0.0049 \pm 0.0010$ & $0.0045 \pm 0.0004$ & $0.704 \pm 0.013$ 
		&- & - & -  \\
		Pop & $0.0297 \pm 0.0005$ & $0.0137 \pm 0.0005$ & $0.904 \pm 0.018$ 
		& $0.0360 \pm 0.0020$ & $0.0171 \pm 0.0008$ & $0.923 \pm 0.023$  \\ 
		BPR & $0.0287 \pm 0.0026$ & $0.0133 \pm 0.0009$ & $0.892 \pm 0.023$ 
		&$0.1200 \pm 0.0036$ & $0.0415 \pm 0.0009$ & $1.387 \pm 0.028$   \\ 
		\midrule
		ULBPR & $0.0505 \pm 0.0021$ & $0.0236 \pm 0.0005$ & $1.037 \pm 0.013$ 
		&$0.1321 \pm 0.0037$ & $0.0446 \pm 0.0009$ & $1.466 \pm 0.021$  \\ 
		ULRMF & $0.0524 \pm 0.0028$ & $0.0237 \pm 0.0006$ & $1.043 \pm 0.015$ 
		&$0.1155 \pm 0.0029$ & $0.0344 \pm 0.0010$ & $1.292 \pm 0.016$  \\ 
		CausE & $0.0797 \pm 0.0031$ & $\textbf{0.0252} \pm 0.0006$ & $1.132 \pm 0.021$ 
		&$\textbf{0.1692} \pm 0.0030$ & $0.0432 \pm 0.0007$ & $\textbf{1.681} \pm 0.017$ \\
		DLTO &$0.0825 \pm 0.0023$ & $0.0226 \pm 0.0007$ & $1.172 \pm 0.019$ 
		&$0.1476 \pm 0.0037$ & $\textbf{0.0452} \pm 0.0010$ & $1.573 \pm 0.025$  \\ 
		BLCE &$0.0374 \pm 0.0025$ & $0.0227 \pm 0.0008$ & $0.990 \pm 0.015$ 
		&$0.1111 \pm 0.0034$ & $0.0449 \pm 0.0010$ & $1.394 \pm 0.025$  \\ 
		DLCE &$\textbf{0.0826} \pm 0.0026$ & $0.0230 \pm 0.0007$ & $\textbf{1.175} \pm 0.020$ 
		&$0.1414 \pm 0.0033$ & $0.0448 \pm 0.0008$ & $1.528 \pm 0.024$  \\
		\bottomrule
	\end{tabular}
\end{table*}

\begin{table*}[htbp]
	\small
	\caption{Performance comparison in product-level granularity of items.
		The best results are highlighted in bold.}
	\label{tab:comparison_product}
	\centering
	\begin{tabular}{lcccccc}
		\toprule
		& \multicolumn{3}{c}{Product-Original} & \multicolumn{3}{c}{Product-Personalized ($\beta = 2.0$)}\\
		\cmidrule(lr){2-4} \cmidrule(lr){5-7} 
		& CP@10 & CP@100 & CDCG & CP@10 & CP@100 & CDCG \\
		\midrule
		Random & $0.0017 \pm 0.0002$ & $0.0017 \pm 0.0001$ & $1.638 \pm 0.008$ 
		&- & - & -  \\
		Pop & $0.0445 \pm 0.0015$ & $0.0198 \pm 0.0002$ & $2.102 \pm 0.012$
		&$0.0572 \pm 0.0032$ & $0.0253 \pm 0.0006$ & $2.264 \pm 0.020$  \\ 
		BPR & $0.0430 \pm 0.0018$ & $0.0203 \pm 0.0003$ & $2.127 \pm 0.014$ 
		&$0.2040 \pm 0.0028$ & $0.0775 \pm 0.0004$ & $3.288 \pm 0.022$  \\
		\midrule
		ULBPR &$0.0604 \pm 0.0019$ & $0.0262 \pm 0.0003$ & $2.253 \pm 0.011$
		&$0.2627 \pm 0.0034$ & $0.0828 \pm 0.0007$ & $3.538 \pm 0.021$  \\ 
		ULRMF &$0.0648 \pm 0.0019$ & $0.0285 \pm 0.0003$ & $2.253 \pm 0.011$
		&$0.2456 \pm 0.0034$ & $0.0769 \pm 0.0006$ & $3.339 \pm 0.019$  \\
		CausE &$0.0592 \pm 0.0021$ & $0.0268 \pm 0.0002$ & $2.285 \pm 0.012$
		&$0.2598 \pm 0.0028$ & $0.0671 \pm 0.0005$ & $3.375 \pm 0.018$  \\ 
		DLTO &$0.0873 \pm 0.0010$ & $0.0310 \pm 0.0002$ & $2.402 \pm 0.009$ 
		&$\textbf{0.2708} \pm 0.0041$ & $\textbf{0.0844} \pm 0.0006$ & $\textbf{3.625} \pm 0.024$  \\ 
		BLCE &$0.0518 \pm 0.0011$ & $0.0302 \pm 0.0002$ & $2.246 \pm 0.012$
		&$0.1980 \pm 0.0029$ & $0.0810 \pm 0.0006$ & $3.310 \pm 0.020$  \\ 
		DLCE &$\textbf{0.0899} \pm 0.0012$ & $\textbf{0.0313} \pm 0.0003$ & $\textbf{2.414} \pm 0.010$ 
		&$0.2671 \pm 0.0029$ & $0.0843 \pm 0.0006$ & $3.586 \pm 0.023$  \\
		\bottomrule
	\end{tabular}
\end{table*}

\begin{table*}[htbp]
	\small
	\caption{Performance comparison in CAR. The smaller is the better in this metric.}
	\label{tab:comparison_car}
	\centering
	\begin{tabular}{lc}
		\toprule
		& Category-Personalized ($\beta = 2.0$)\\
		& CAR \\
		\midrule
		DLTO &$0.770 \pm 0.026$ \\
		BLCE &$0.694 \pm 0.049$ \\
		DLCE &$\textbf{0.649} \pm 0.040$ \\
		\bottomrule
	\end{tabular}
\end{table*}

\subsection{Dependence on the propensity capping}
\label{subsec:sensitivity_hyperparameters}
\begin{figure*}[htbp]
	\begin{center}
		\subfigure[Category-Original.]{\includegraphics[width=0.245\textwidth]{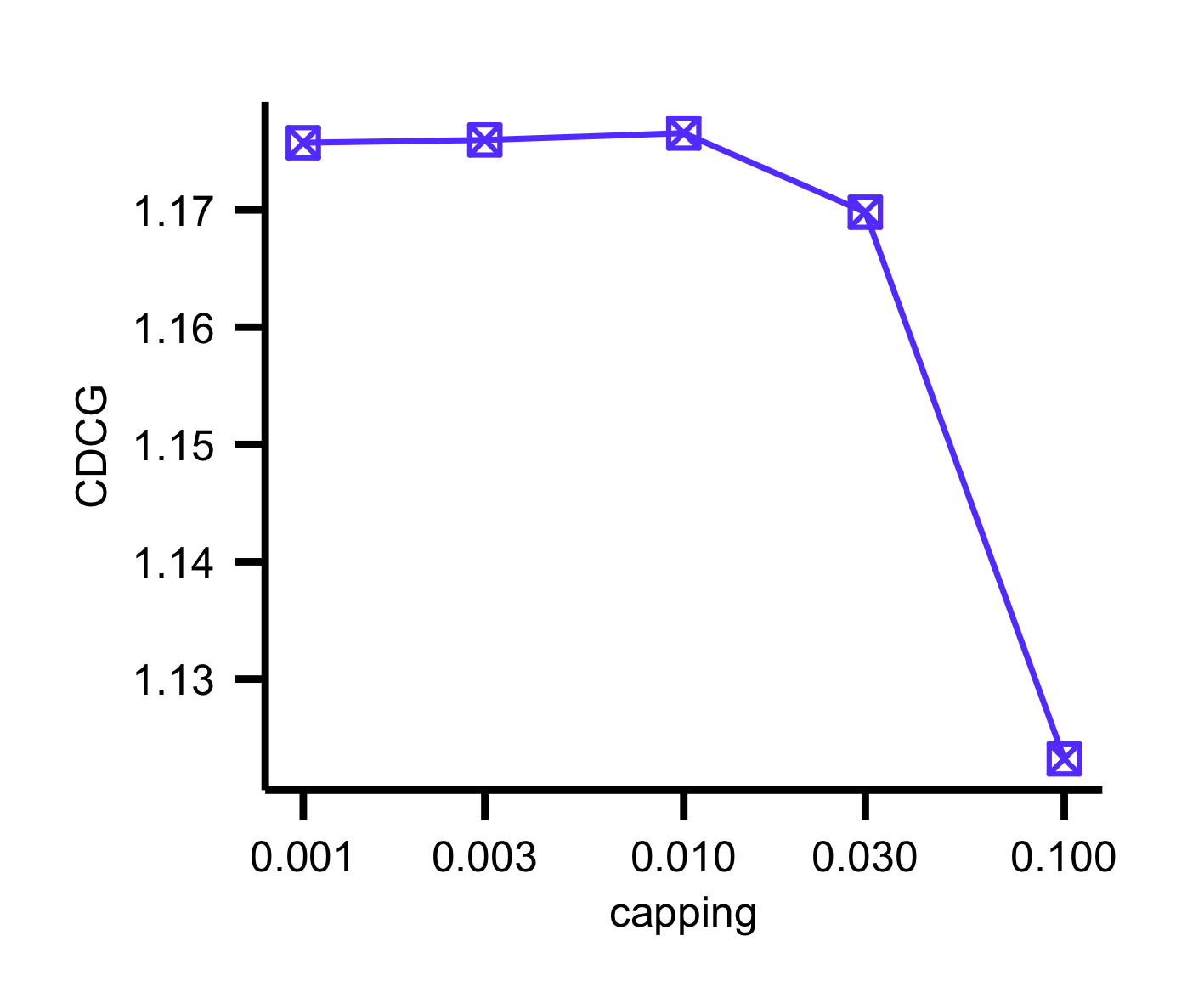}}
		\subfigure[Category-Personalized.]{\includegraphics[width=0.245\textwidth]{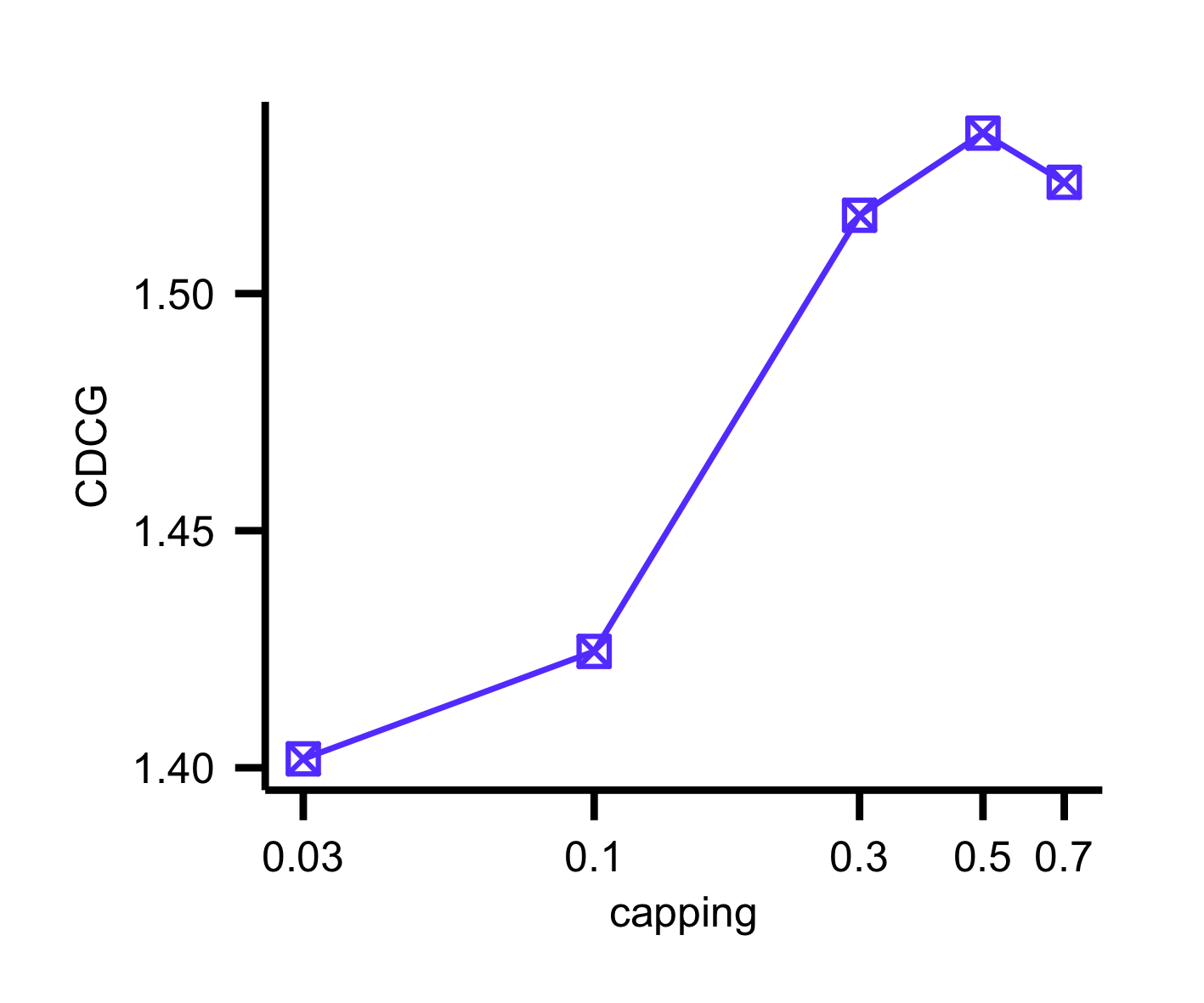}}
		\subfigure[Product-Original.]{\includegraphics[width=0.245\textwidth]{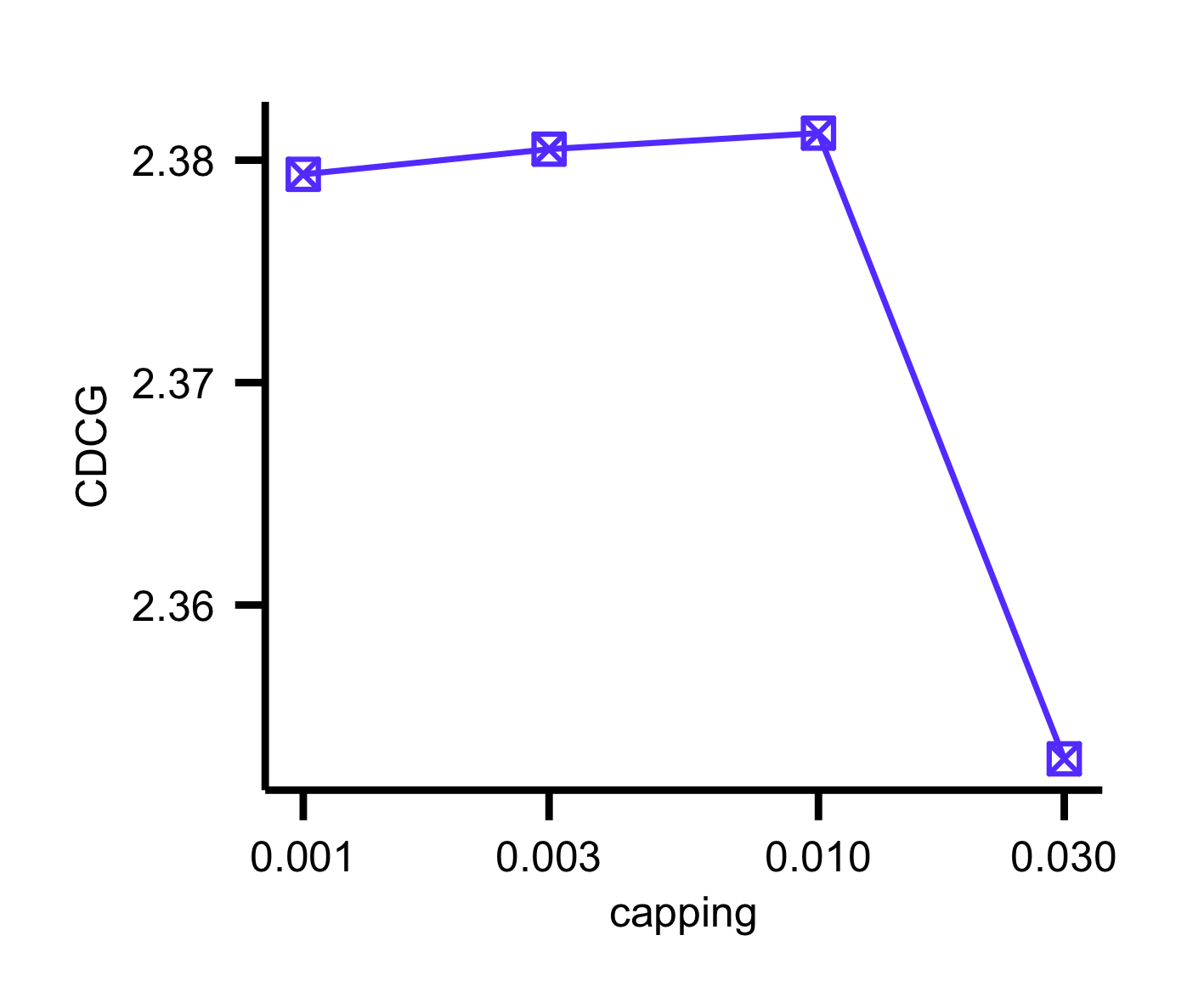}}
		\subfigure[Product-Personalized.]{\includegraphics[width=0.245\textwidth]{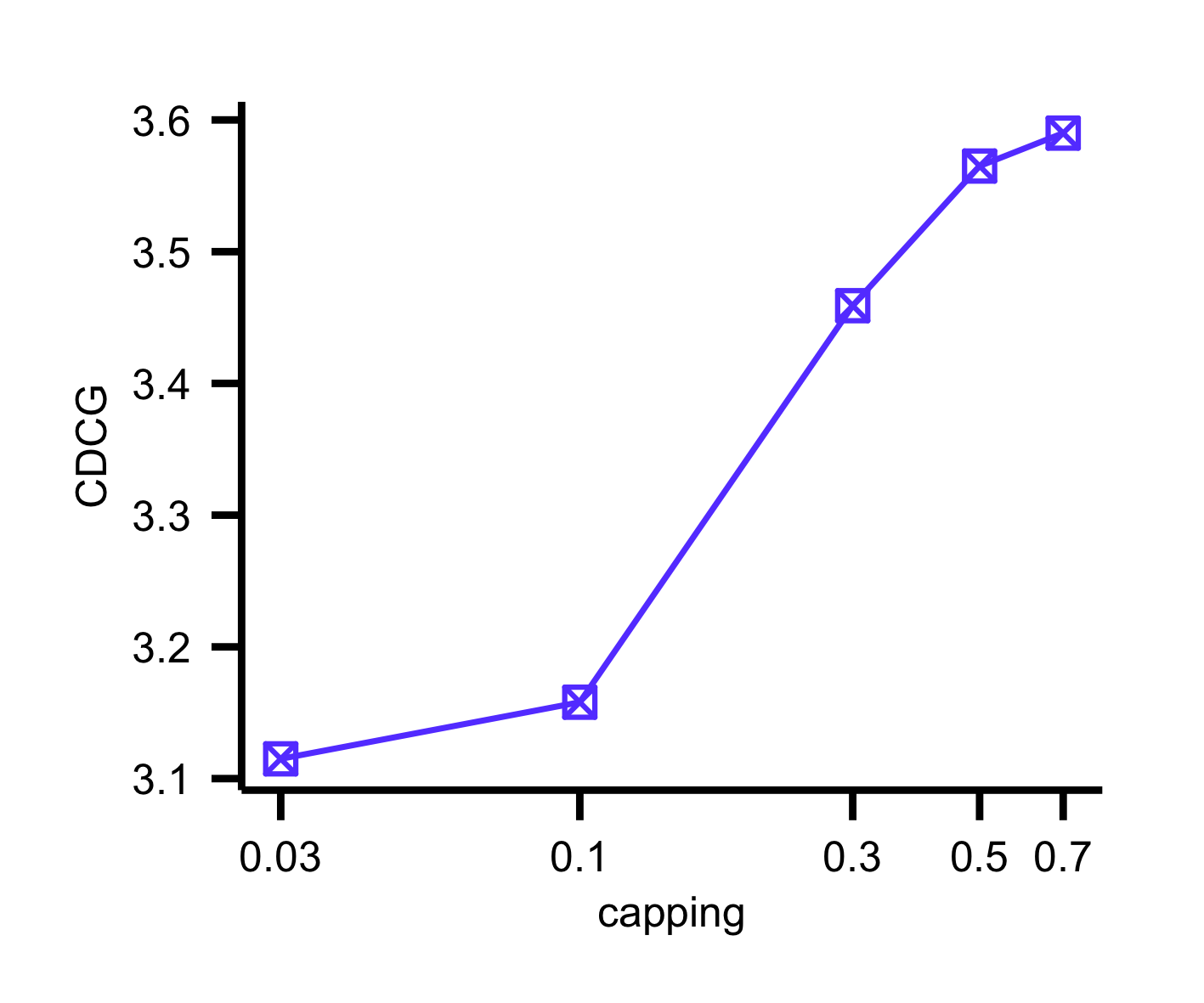}}
		\caption{Dependence on the capping thresholds.}
		\label{fig:capping}
	\end{center}
\end{figure*}

\begin{figure*}[htbp]
	\begin{center}
		\subfigure[Category-Personalized.]{\includegraphics[width=0.245\textwidth]{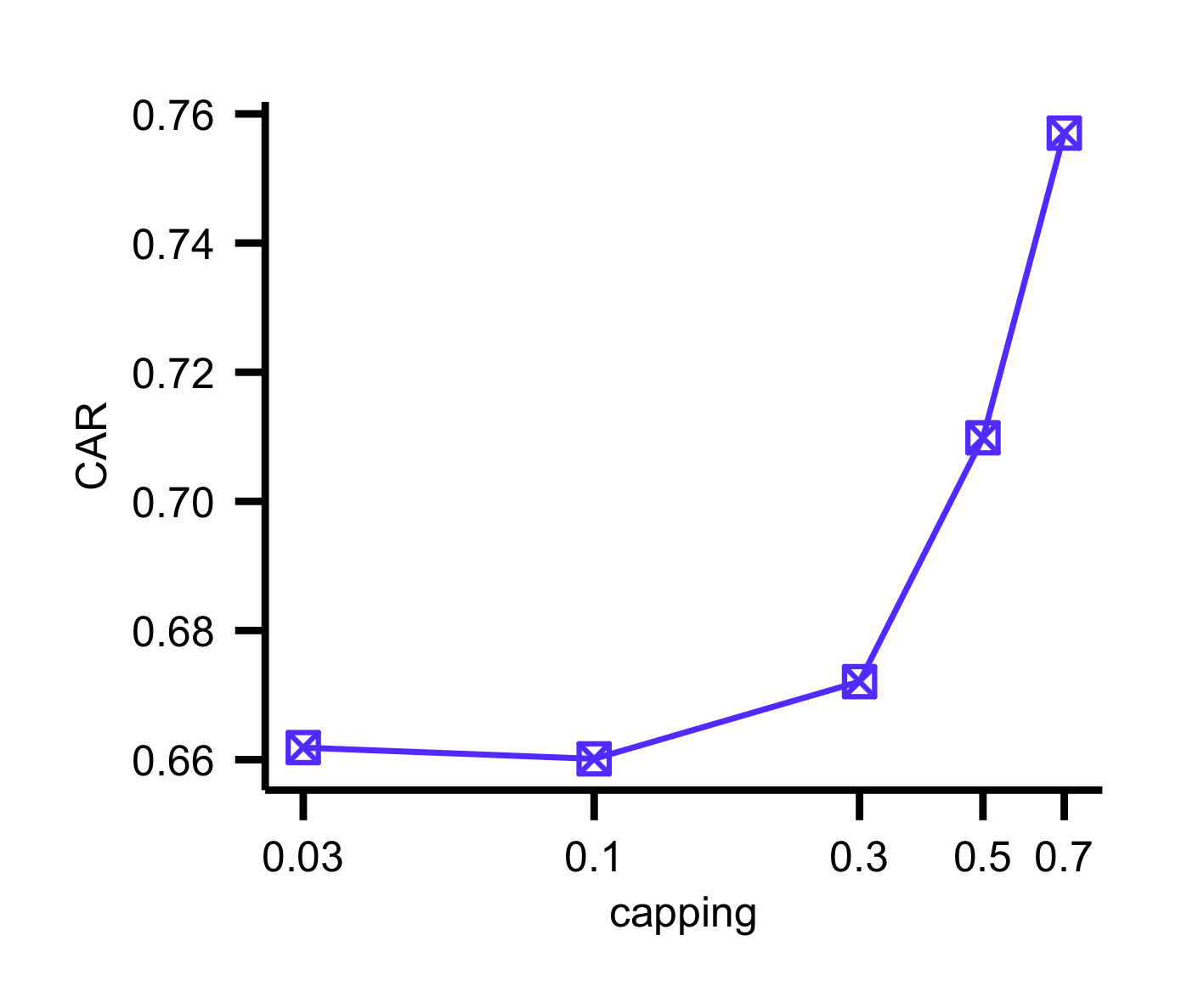}}
		\caption{Dependence on the capping thresholds for CAR. The smaller is the better.}
		\label{fig:capping_car}
	\end{center}
\end{figure*}

We investigate the dependence on the capping threshold $\chi$.
As shown in Fig. \ref{fig:capping}, the performance improves with appropriate $\chi$, showing the importance of the propensity capping.
The best values are different for each dataset and they tend to be larger for the personalized settings.
This is probably because the distributions of propensities are heavily skewed in personalized settings as shown in Fig. \ref{fig:propensity_dist}.
We also investigate the capping dependence in CAR metric (Fig. \ref{fig:capping_car}).
The optimal cappings are around 0.1 and are less than the values for CDCG.

\subsection{Reliability of unbiased estimators for evaluation}
\label{subsec:reliability_unbiased_estimators}
Since the ground truth of causal effect $\tau_{ui}$ is unobservable, the ranking performance shown in Tables \ref{tab:comparison_category} and \ref{tab:comparison_product} cannot be directly obtained in practice.
This means that we need to resort to its estimator for the performance testing and the hyperparameter tuning.
Therefore, we investigate the reliability of our IPS-based unbiased estimators of the metrics introduced in Subsection \ref{subsec:unbiased_estimator}.
For $n_{test}=10$ test datasets, we calculate mean absolute error (MAE) of the estimators with different capping.
The results are shown in Table \ref{tab:valid_estimator}.
In the original setting where the unevenness of propensity is mild, MAEs are small enough for coarse investigation of the performance difference among methods or roughly tuning hyperparameters.
In the personalized setting where the uneveness of propensity is severe (see Fig. \ref{fig:propensity_dist}), MAEs are large.
We additionally investigated the MAEs for the estimate of CAR.
As shown in Table \ref{tab:car_estimator}, the MAEs were reduced with the proper capping value.
Our learning methods optimize for CAR, hence this reliable estimate even for the severe unevenness of propensity leads to the successful results in the personalized settings.
However, the estimators for other metrics might need to be improved.
The estimators can be enhanced, for example, by applying a doubly robust method \cite{Wang19} that employs both IPS weighting and predictors of potential outcomes.
This could be considered as future work.

\begin{table*}[htbp]
	\small
	\caption{MAEs of unbiased estimators at varied capping thresholds. The lowest MAEs for each metric-method pair are in bold.}
	\label{tab:valid_estimator}
	\centering
	\begin{tabular}{llcccccccccc}
		\toprule
		& & \multicolumn{5}{c}{Category-Original} & \multicolumn{5}{c}{Category-Personalized}\\
		\cmidrule(lr){3-7} \cmidrule(lr){8-12} 
		& & $\chi$=0.0 & 0.003 & 0.01 & 0.03 & 0.1 & $\chi$=0.0 & 0.01 & 0.03 & 0.1 & 0.3\\
		\midrule
		CP@10 & Pop & \textbf{0.0129} & \textbf{0.0129} & \textbf{0.0129} & \textbf{0.0129} & 0.0130
		& \textbf{0.1527} & \textbf{0.1527} & \textbf{0.1527} & 0.1528 & 0.1530 \\ 
		& BPR & \textbf{0.0110} & \textbf{0.0110} & \textbf{0.0110} & 0.0113 & 0.0131
		& \textbf{0.1523} & \textbf{0.1523} & \textbf{0.1523} & 0.1524 & 0.1524 \\ 
		& ULBPR & \textbf{0.0040} & \textbf{0.0040} & \textbf{0.0040} & 0.0044 & 0.0069
		& 0.1343 & 0.1343 & 0.1343 & 0.1342 & \textbf{0.1320} \\ 
		& DLCE &  0.0063 & 0.0063 & 0.0063 & \textbf{0.0061} & 0.0150
		& 0.1324 & 0.1324 & 0.1324 & 0.1324 & \textbf{0.1308} \\
		\midrule
		CP@100 & Pop & 0.0018 & 0.0018 & \textbf{0.0016} & 0.0018 & 0.0021
		&  0.0544 & 0.0544 & 0.0545 & 0.0546 & \textbf{0.0540} \\ 
		& BPR & 0.0018 & 0.0016 & 0.0012 & \textbf{0.0010} & 0.0022
		& 0.0535 & 0.0535 & 0.0535 & 0.0535 & \textbf{0.0522} \\ 
		& ULBPR & \textbf{0.0017} & \textbf{0.0017} & 0.0018 & 0.0020 & 0.0019
		& 0.0569 & 0.0569 & 0.0570 & 0.0568 & \textbf{0.0549} \\ 
		& DLCE &  \textbf{0.0016} & \textbf{0.0016} & \textbf{0.0016} & 0.0018 & 0.0019 
		& 0.0604 & 0.0604 & 0.0604 & 0.0603 & \textbf{0.0587} \\
		\midrule
		CDCG & Pop & 0.076 & 0.059 & \textbf{0.042} & 0.081 & 0.282
		&  2.081 & 2.076 & 2.027 & 1.914 & \textbf{1.717} \\ 
		& BPR & 0.072 & 0.061 & \textbf{0.048} & 0.090 & 0.303
		& 2.005 & 2.000 & 1.949 & 1.836 & \textbf{1.638} \\ 
		& ULBPR & 0.058 & 0.050 & \textbf{0.044} & 0.091 & 0.283
		& 1.830 & 1.824 & 1.766 & 1.652 & \textbf{1.461} \\ 
		& DLCE &  0.067 & \textbf{0.065} & 0.066 & 0.125 & 0.398 
		& 1.911 & 1.905 & 1.852 & 1.739 & \textbf{1.539} \\ 
		\bottomrule
	\end{tabular}
\end{table*}

\begin{table*}[htbp]
	\small
	\caption{MAEs of unbiased estimators for the causal average rank. }
	\label{tab:car_estimator}
	\centering
	\begin{tabular}{llccccc}
		\toprule
		& & \multicolumn{5}{c}{Category-Personalized}\\
		\cmidrule(lr){3-7} 
		& & $\chi$=0.0 & 0.01 & 0.03 & 0.1 & 0.3\\
		\midrule
		CAR & POP & 0.805 & 0.759 & 0.399 & \textbf{0.075} & 0.623\\
		    & DLCE & 0.866 & 0.834 & 0.537 & \textbf{0.072} & 0.640\\
		\bottomrule
	\end{tabular}
\end{table*}

\subsection{Influence of the unevenness of propensity on performance}
\label{subsec:unevenness_propensity}
\begin{figure}[htbp]
	\begin{center}
		\subfigure[Category-Personalized.]{\includegraphics[width=0.4\textwidth]{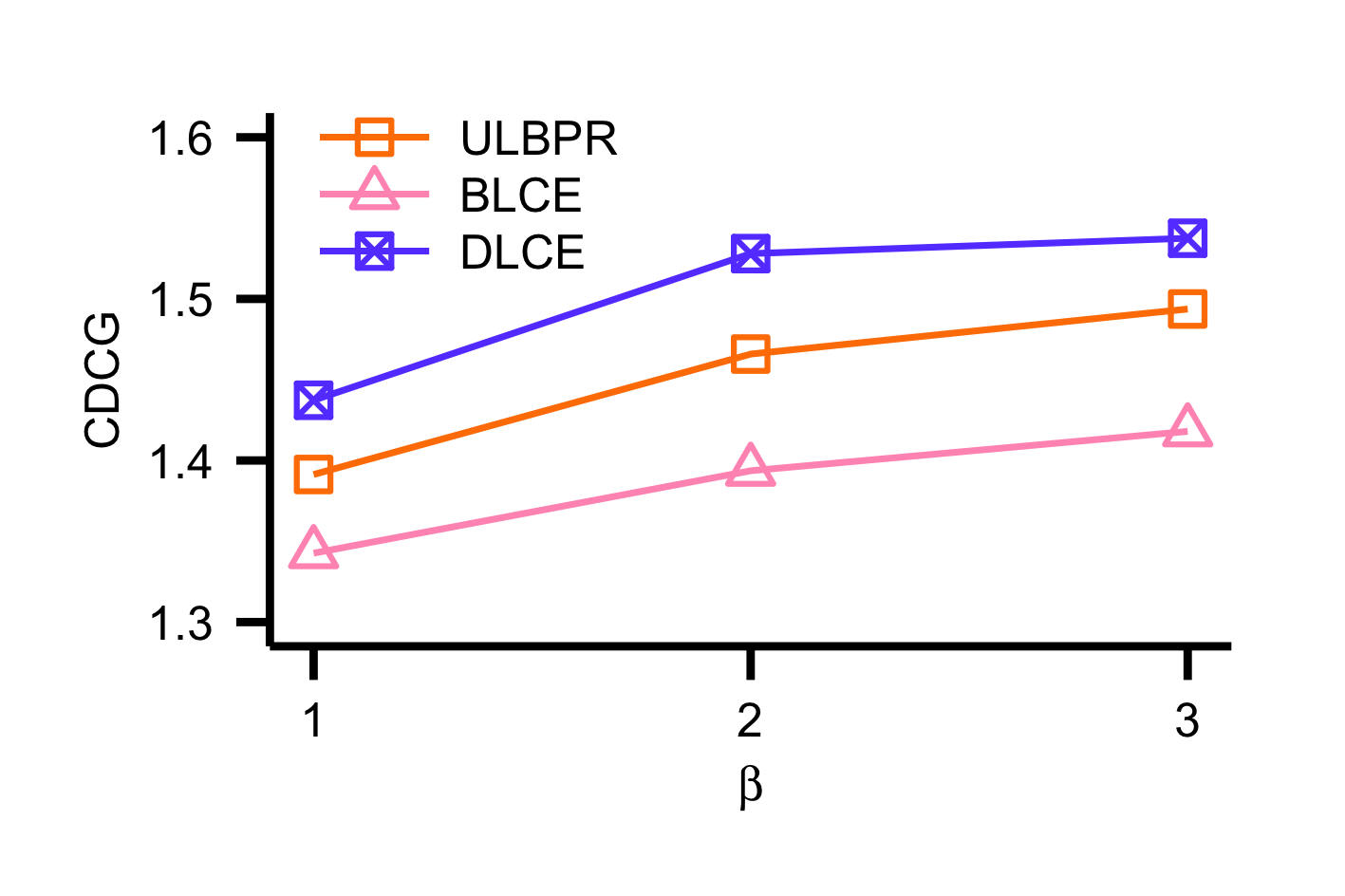}}
		\subfigure[Product-Personalized.]{\includegraphics[width=0.4\textwidth]{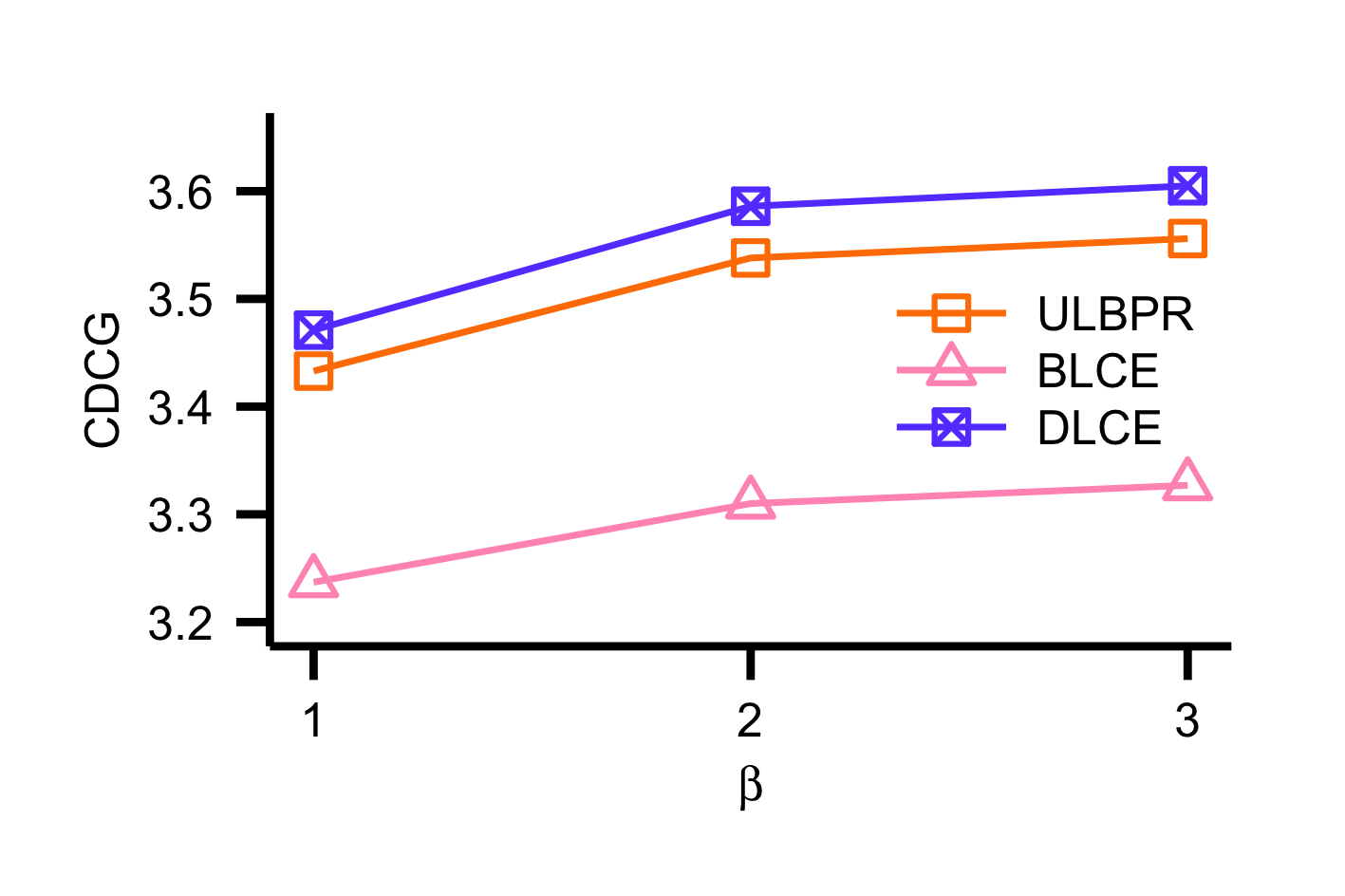}}
		\caption{Performance under the varied unevenness of propensity.
		The parameter $\beta$ determines the unevenness as expressed in Eq. (\ref{eq:gen_propensity})}
		\label{fig:unevenProp}
	\end{center}
\end{figure}
We vary the unevenness of the propensity (i.e., the severity of bias) and evaluate the performance.
Fig. \ref{fig:unevenProp} shows the result.
A higher value of $\beta$ makes propensities more skewed to the top position of item ranking in terms of the user's preference.
Our DLCE outperforms other methods for varied unevenness.

\subsection{Robustness of DLCE to the misspecified propensity}
\begin{figure*}[htbp]
	\begin{center}
		\subfigure[Category-Original.]{\includegraphics[width=0.245\textwidth]{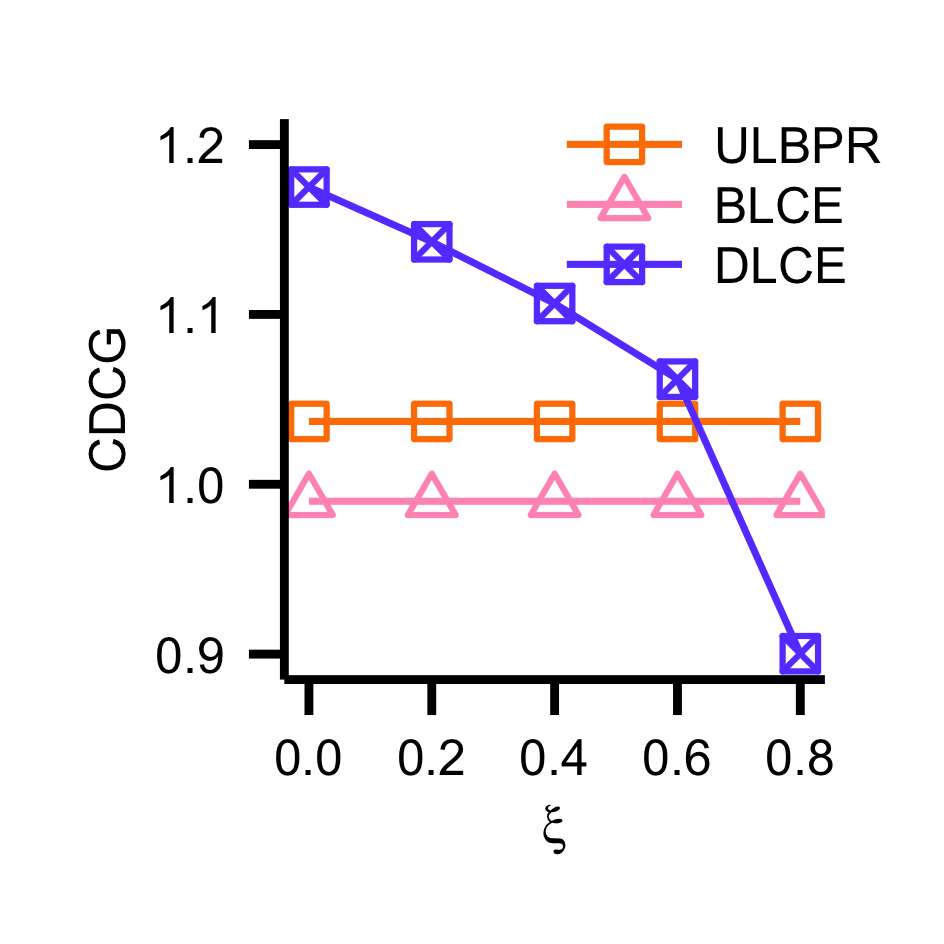}}
		\subfigure[Category-Personalized.]{\includegraphics[width=0.245\textwidth]{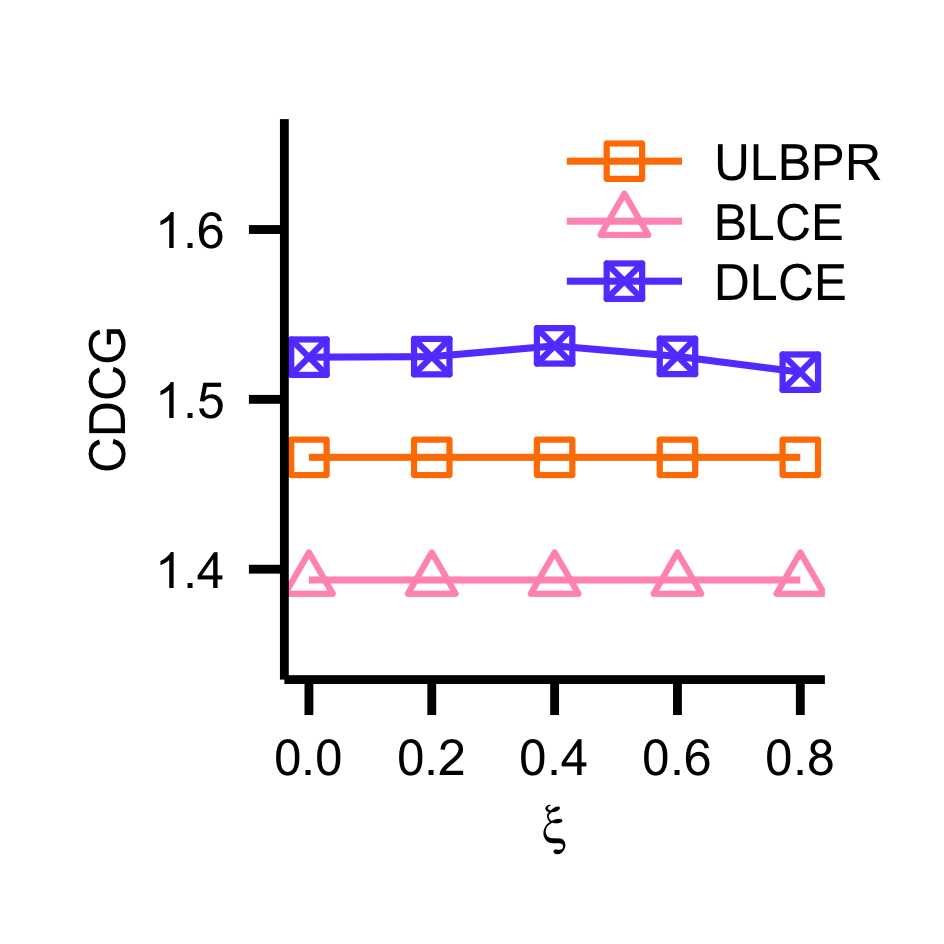}}
		\subfigure[Product-Original.]{\includegraphics[width=0.245\textwidth]{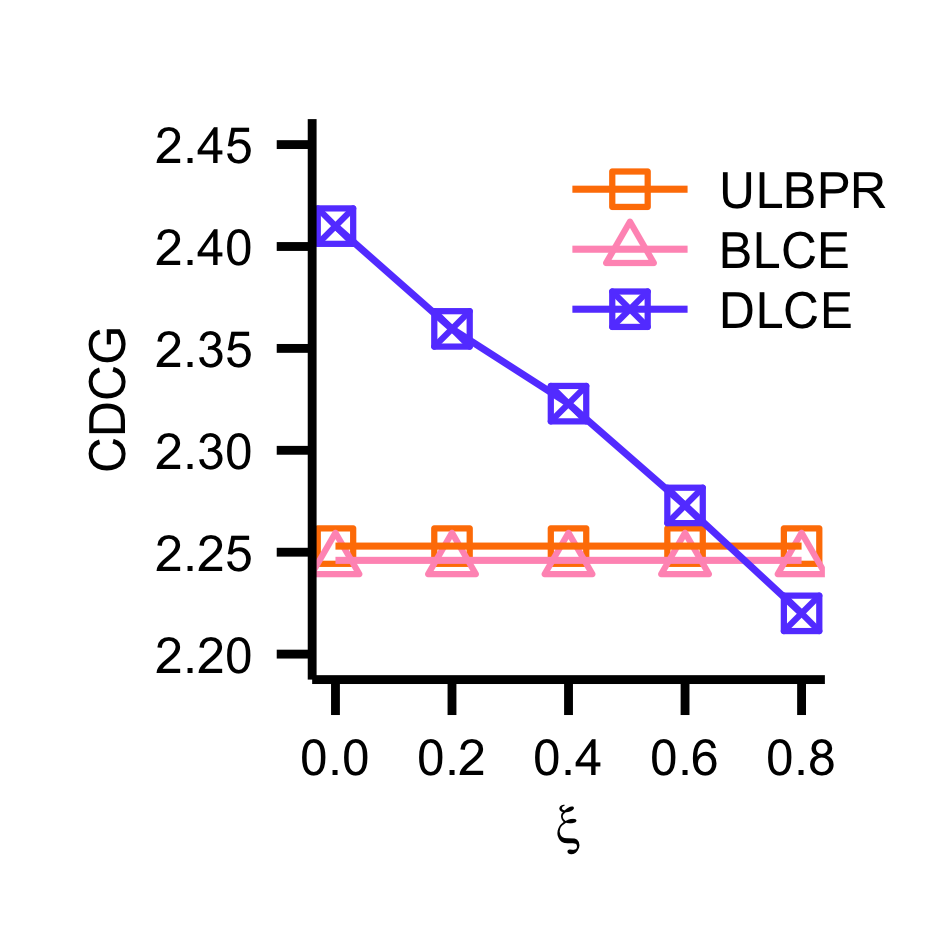}}
		\caption{Robustness to misspecified propensity. 
			$\xi$ adjusts the severity of missspecification as expressed in Eq. (\ref{eq:misIPS})}
		\label{fig:misIPS}
	\end{center}
\end{figure*}
Thus far, all experiments assumed that the true propensity is known (i.e., recorded as in \cite{Lefortier16}.)
However, if the true propensity is not accessible, it needs to be estimated and is subjected to misspecification.
Here, we evaluate the robustness of DLCE to the misspecification.
We simulate the misspecification of propensity so that the log-odd of the propensity shifts to its mean.
\begin{equation}
\label{eq:misIPS}
\log \left(\frac{P'_{ui}}{1-P'_{ui}} \right) = (1-\xi)\log \left(\frac{P_{ui}}{1-P_{ui}} \right) + \xi \frac{1}{UI} \sum_u \sum_i \log \left(\frac{P_{ui}}{1-P_{ui}} \right),
\end{equation}
where $P'_{ui}$ is the misspecified propensity and $\xi$ is the parameter to adjust the severity of misspecification.
The reason of this expression is that the logistic regression is often used for propensity estimate and the prediction of logistic regression becomes close to the mean of log-odds by the regularization.

Fig. \ref{fig:misIPS} shows the performance of DLCE with varied levels of misspecification.
ULBPR and BLCE, biased learning methods, do not use the propensity and are not affected by the misspecification.
The result shows that DLCE can substantially improves upon BLCE and ULBPR for a wide range of misspecification severity $\xi$.
We set $\chi=0.01$ and $0.5$ for the original and the personalized settings, respectively.
Relatively small influence of the misspecification on the personalized setting might be due to the large capping threshold, and increasing the capping threshold for original setting could alleviate the degrade from misspecification.
Overall, we conclude that DLCE is effective even if the true propensity is unknown and needs to be estimated.

\section{Conclusions}
In this paper, we proposed an unbiased learning framework for the causal effect of recommendation.
Based on the IPS weighting technique, the proposed framework first constructs unbiased estimators for ranking metrics.
Then, it conducts ERM on the estimators with propensity capping that reduce variance under finite training samples.
Based on the framework, we developed an efficient debiased learning method with SGD for the causal average rank.
We empirically showed that the proposed method outperforms baselines in various settings and is robust to the severity of recommendation bias and the misspecification of propensity.

This study opens several directions for future research.
First, the proposed framework enables future studies to develop learning methods to optimize other metrics such as causal DCG.
Several extensions \cite{Hu19, Agarwal19} for other metrics followed after early work \cite{Joachims17} of unbiased learning for selection bias.
Second, the IPS-based unbiased estimator can be further enhanced with a doubly robust method \cite{Wang19} or variance reduction methods \cite{Swaminathan15a, Swaminathan15b}.
Third, the proposed method, which is meant for only one type of recommendation, can be extended to multiple types of recommendations where items are recommended by e-mail, pop-up after log-in, etc.


\providecommand*\hyphen{-}\providecommand*\quotation{’}

\end{document}